\documentclass[]{ceurart}

\usepackage{listings}
\usepackage{amsmath,amssymb}

\usepackage{subcaption}

\begin{document}

\copyrightyear{2023}
\copyrightclause{Copyright for this paper by its authors.
  Use permitted under Creative Commons License Attribution 4.0
  International (CC BY 4.0).}

\conference{In A. Martin, K. Hinkelmann, H.-G. Fill, A. Gerber, D. Lenat, R. Stolle, F. van Harmelen (Eds.), 
Proceedings of the AAAI 2023 Spring Symposium on Challenges Requiring the Combination of Machine Learning and Knowledge Engineering (AAAI-MAKE 2023), Hyatt Regency, San Francisco Airport, California, USA, March 27-29, 2023.}

\title{The Contribution of Knowledge in Visiolinguistic Learning: A Survey on Tasks and Challenges}

\author[1]{Maria Lymperaiou}[%
orcid=0000-0001-9442-4186,
email=marialymp@islab.ntua.gr]
\address[1]{AILS Laboratory,
School of Electrical and Computer Engineering,
National Technical University of Athens}

\author[1]{Giorgos Stamou}[orcid= 0000-0003-1210-9874, email=gstam@cs.ntua.gr]

\begin{abstract}
Recent advancements in visiolinguistic (VL) learning have allowed the development of multiple models and techniques that offer several impressive implementations, able to currently resolve a variety of tasks that require the collaboration of vision and language. Current datasets used for VL pre-training only contain a limited amount of visual and linguistic knowledge, thus significantly limiting the generalization capabilities of many VL models. External knowledge sources such as knowledge graphs (KGs) and Large Language Models (LLMs) are able to cover such generalization gaps by filling in missing knowledge, resulting in the emergence of hybrid architectures. In the current survey, we analyze tasks that have benefited from such hybrid approaches. Moreover, we categorize existing knowledge sources and types, proceeding to discussion regarding the KG vs LLM dilemma and its potential impact to future hybrid approaches.
\end{abstract}


\maketitle

\section{Introduction}
\label{sec:intro}
Visiolinguistic (VL) learning has been one of the fastest evolving fields of artificial intelligence, especially after the emergence of the Transformer \cite{transformer}, which enabled a variety of powerful architectures. Popular VL tasks such as Visual Question Answering (VQA)  \cite{vqa}, Visual Reasoning (VR) \cite{visual_reasoning}, Visual Commonsense Reasoning (VCR) \cite{vcr}, Visual Entailment (VE) \cite{entailment}, Image Captioning (IC) \cite{captioning}, Image-Text Retrieval (ITR) and inversely Text-Image Retrieval (TIR) \cite{retrieval}, Visual-Language Navigation (VLN) \cite{navigation}, Visual Storytelling (VIST) and Visual Dialog (VD) \cite{dialog-gen} have been significantly benefited from recent transformer-based advancements which follow the \textit{pre-train fine-tune} learning framework. \textit{Pre-training} is responsible of fusing generic information regarding  visual and linguistic patterns, as well as how those two modalities interact, based on information present in large-scale datasets. Self-supervised objective functions are employed to help the VL model learn interdependencies between vision and language during pre-training. For example, masking out words from image captions enforces learning how to fill them based on visual cues; reversely, image regions can be masked out, with language guiding their reconstruction.
Task-specific \textit{fine-tuning} steps upon this basic understanding of vision and language, by refining the neural weights of the trained model to adapt to each specific task at a time, upon which the final evaluation is performed.

\begin{figure}[h!]
     \centering
     \begin{subfigure}[b]{0.2\textwidth}
         \centering
         \includegraphics[width=\textwidth, height=2.8cm]{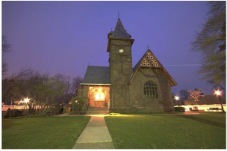}
         \caption*{Q: What days might I most commonly go to this building? A: Sundays.}
         \label{fig:a}
     \end{subfigure}
     \hfill
     \begin{subfigure}[b]{0.2\textwidth}
         \centering
         \includegraphics[width=\textwidth, height=2.8cm]{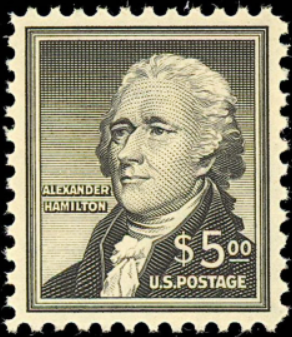}
         \caption*{Q: In which continent was the person in the image born? A: North America.}
         \label{fig:b}
     \end{subfigure}
     \hfill
     \begin{subfigure}[b]{0.2\textwidth}
         \centering
         \includegraphics[width=\textwidth, height=2.8cm]{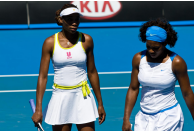}
         \caption*{Q: Who among the people
in the image is the eldest? A: Person in the left.}
         \label{fig:c}
     \end{subfigure}
         \hfill
     \begin{subfigure}[b]{0.2\textwidth}
         \centering
         \includegraphics[width=\textwidth, height=2.8cm]{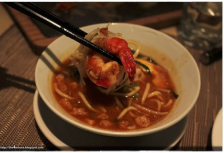}
         \caption*{Q: What is the name of the object used to eat this food? A: Chopsticks.}
         \label{fig:d}
     \end{subfigure}
        \caption{External knowledge is required to answer these visual questions \cite{okvqa, vqa5, vqa18}.}
        \label{fig:examples}
\end{figure}

Despite the rich VL knowledge acquired during this process, current transformer-based VL models \cite{lu2019vilbert, li2020oscar, flava, kim2021vilt, simvlm, clip, soho, florence, align} lack generalization to several concepts and scenarios that require \textbf{commonsense
knowledge}, or knowledge of \textbf{abstract entities}, \textbf{facts} and \textbf{real-world events}. Of course, this is somehow expected, since neither pre-training nor fine-tuning VL datasets contain or demand perceiving concepts beyond visual descriptions. Figure \ref{fig:examples} presents some examples of this claim: questions (\textit{Q}) about the image (\textit{I})
require some knowledge beyond the visual domain, so that the correct answer (\textit{A}) can be inferred.

The first image of Figure \ref{fig:examples} requires knowledge about\textbf{ human culture and history} \cite{okvqa}, to combine with \textbf{visual} information: the object in the image is a \textit{church}, and \textit{people usually go to the church on Sundays}. The second image \cite{vqa18} requires one more reasoning step, since it is not only required to detect that this is a postage stamp containing the photo of a \textit{person} (\textbf{visual} information), but also who this person is. Knowledge about \textbf{named entities} recognizes this person as \textit{Alexander Hamilton}. Further \textbf{factual} knowledge provides that  \textit{Alexander Hamilton was born in todays Saint Kitts and Nevis} and \textit{Saint Kitts and Nevis is in North America}. The combination of these two facts derives the final answer \textit{Alexander Hamilton was born in North America}. The third image \cite{vqa5} requires the \textbf{visual} extraction of the two \textit{people} present in it. Then, \textbf{named entities} knowledge assigns the identities \textit{Serena Williams} and \textit{Venus Williams} to these two people. Their \textit{age} is provided as a combination of \textbf{named entities} and \textbf{factual} knowledge, yielding the \textbf{comparative} knowledge fact that \textit{Serena Williams is older than Venus Williams}. Finally, \textbf{spatial} knowledge derives that \textit{Serena Williams is the person in the left}. The overall combination of \textbf{named entities}, \textbf{comparative} and \textbf{spatial} knowledge returns the final answer \textit{The person in the left}. It becomes obvious that answering these question requires more knowledge from external sources, which is extracted and combined to infer an answer.

Thus, the incorporation of external knowledge in earlier or later stages of the \textit{pre-training/fine-tuning} process is necessary to enhance the capabilities of VL models, so that they are able to respond to more real-world scenarios.  Such knowledge is typically represented using entities, relationships and semantic descriptions \cite{kg1}
stored in structured Knowledge Graphs (KGs) \cite{wordnet, conceptnet, wikidata, dbpedia}. Language Models (LMs) such as BERT \cite{bert} have been proven capable of storing relational knowledge learned from linguistic data during pre-training, introducing the LM-as-KB scenario \cite{lmkb1}. This knowledge can then be retrieved by constructing queries as fill-the-blank statements, which the LM is tasked to complete. Further works validate the abilities of LMs for world-knowledge storage and retrieval, while showcasing their scaling capacity according to the number of parameters \cite{lmkb3}. There are some prerequisites for LM to successfully serve as knowledge bases; accessing the data similarly to KG querying, updating outdated facts while trespassing the risk of catastrophic forgetting, unlocking their rather obscure reasoning capabilities and measuring the degree of their interpretability and explainability are still open challenges \cite{lmkb2}. 
More recently, the impressive results of Large Language Models (LLMs) \cite{gpt3, chinchilla, palm, glam, gopher} in various linguistic tasks greatly inspire their possible usage as rich and simultaneously vast knowledge bases (KBs) to aid VL learning. 

Prior surveys in VL learning \cite{survey0, survey1, survey2, survey3, survey4, survey5, survey6} do not focus on the collaboration between knowledge and deep learning VL models. An exhaustive presentation of the knowledge-enhanced VL (KVL) topic was presented in \cite{kvl-survey} for the first time.
In the current survey paper, we focus on state-of-the-art endeavors involving transformer models for the VL representation, leading to  \textit{hybrid} approaches when combined with external knowledge. Finally, we discuss around potential trends regarding the external knowledge assisting VL models and how it is expected to affect future applications in the field.

\section{Knowledge and Reasoning}
\subsection{Types of external knowledge}
External knowledge sources are divided in two main categories, \textit{explicit} and \textit{implicit} \cite{kvl-survey}. They are both capable of providing \textbf{factual}, \textbf{commonsense}, \textbf{temporal}, \textbf{lexical} or other knowledge senses \cite{commonsense-dimensions} missing from pre-trained VL models. The type of the external knowledge source used significantly defines the way of retrieving and harnessing knowledge for VL models.
\paragraph{Explicit knowledge} refers to the knowledge stored in KGs in a structured format. Such knowledge is symbolically represented in the form of triplets \textit{(h, r, t)}, which contain entities \textit{(h, t)} and their in-between relationships \textit{r}. Extracting an answer from a KB is a fully transparent process, and the path followed can be deterministically recovered. This is crucial especially when evaluating multi-step and compositional reasoning, so that the factuality of the reasoning path followed is guaranteed. 
Nevertheless, crafting and maintaining KGs requires manual effort or supervision, therefore hindering the automatic extension of such KBs.

Popular open-source knowledge graphs that have contributed to VL learning are ConceptNet \cite{conceptnet}, DBPedia \cite{dbpedia}, Wikidata \cite{wikidata}, YAGO \cite{yago4} and others.
The retrieval of KG facts is primarily based on SPARQL queries, a clear and deterministic query language designed for this purpose. Many open-source knowledge graphs also provide APIs for frequent queries. Due to the constrains SPARQL querying imposes, bridging the gap between natural language user queries and SPARQL has been a useful venture \cite{kg-query}.

Knowledge graph representation learning provides low-dimensional distributed vectors, following the popular strategy of linguistic embeddings \cite{kg1}. This approach allows the application of Deep Learning techniques on KGs, while a better-suited communication between KGs and VL models is established, since all three modalities can be viewed as numerical vectors.

\paragraph{Implicit knowledge} covers unstructured knowledge stored in neural weights, depicting facts and relationships learned during model pre-training. This process allows the integration of multiple and large-scale data sources without the need for human supervision. Existing knowledge can be extended and updated by re-training or fine-tuning the existing neural network (LM/LLM). However, this process is computationally prohibitive for most research institutions, while the factuality, fairness and trustworthiness of learned knowledge and reasoning are questionable, since they significantly depend on the quality of the training data used.

There is an ongoing list of LMs-as-KB, since any language model pre-trained under self-supervised learning objectives can potentially serve as a KB. Retrieving knowledge from LMs is not as straightforward as for KGs, due to the opaque LM structure. Therefore, there are two main ways to access LM knowledge, \textit{indirect access} (\textit{fine-tuning}) and \textit{direct access} (\textit{prompting}).

\textit{Fine-tuning} has been the most concrete way to exploit LM knowledge, even if it does not actually retrieve existing knowledge. Similarly to how fine-tuning works for VL models (Section \ref{sec:intro}), LM fine-tuning refers to adapting neural weights towards a downstream linguistic task by training for a few epochs on a small labelled linguistic dataset, appropriate for the desired task.

\textit{Prompting} obeys to the pre-train, prompt, predict pipeline \cite{prompt-survey}, which allows direct access to information stored in a pre-trained LM, should an appropriate prompt is designed. This is where the difficulty of this approach lies: composing an optimal prompt is an open research topic, and sub-optimal prompt templates may just denote the lower bound of knowledge contained within LMs \cite{prompt-how, prompt-how2}. Prompts in textual format, called \textit{discrete prompts} are quite intuitive to humans, therefore hints on how to craft them can be based on human conversational behavior. Therefore, mining templates from large corpora \cite{prompt-how2}, paraphrasing of existing prompts \cite{prompt-how2, bertese}, fill-blank via language generation \cite{prompt-gen, rl} and others
have been proposed as viable directions. On the other hand, \textit{soft prompts} circumvent interpretability in the sake of efficiency, by directly accessing the LM's embedding space. Prefix-tuning using continuous task-specific vectors in a frozen LM \cite{prefix-tuning}, soft prompting based on discrete prompting initialization \cite{factual-probing} and others have demonstrated promising results. The non-trivial search for prompt-based knowledge retrieval is rewarded with few-shot or even zero-shot reasoning, able to revolutionize KVL models.

\subsection{Reasoning in Knowledge Graphs and Large Language Models}
\label{sec:reasoning}
Reasoning has been one of the milestones regarding the capacity of LLMs to sufficiently perform as KBs. It is regarded as the capability of drawing conclusions and making decisions based on given information, and can be divided in \textit{formal} and \textit{informal}. \textit{Formal} reasoning refers to following a set of rules in a logical and deterministic manner. On the other hand, \textit{informal} reasoning is mostly based on a generic experience and intuition of the world, thus being prone to errors, while however being more flexible \cite{reasoning}. 

KGs tend to approach the \textit{formal} reasoning path, due to their structured nature and the determinism accompanying decision-making. \textit{Informal} reasoning is mostly interconnected with unstructured KBs, which have acquired a more probabilistic sense of the world. LMs have demonstrated some adequate reasoning capabilities, as long as they are large enough, thus belonging to LLMs \cite{emerging}. Nevertheless, the landscape of the full potential of LLM reasoning has not yet been entirely explored. Prompting has been utilized towards unlocking reasoning capabilities of LLMs, encouraging them to reveal their Chain-of-Thought (CoT) instead of merely providing the final answer \cite{cot, cot-reasoning}. CoT has been proven successful towards unveiling hidden reasoning capabilities, either in the few-shot setting \cite{cot-reasoning}, where the LLM is prompted with some exemplars of the desired reasoning, together with an instructive phrase, or in the zero-shot setting \cite{cot}, where the model receives an instructive phrase without exemplars. Another line of work proposes the evaluation of LLMs on downstream tasks exploiting well-crafted datasets, each of which is dedicated on different reasoning senses. To this end, various tests have stressed LLM capabilities on arithmetic \cite{mathqa}, symbolic \cite{cot-reasoning}, commonsense \cite{commonsense-survey}, and other types of reasoning. Overall, the findings occurring from the aforementioned endeavors suggest that indeed, LLMs present \textit{emergent} reasoning capabilities simulating human thinking patterns, though being incapable of tackling complex reasoning challenges. Nevertheless, such evidence cannot conclude whether LLM present \textit{real} reasoning capabilities or if they can just perfectly overfit on the vast information they receive \cite{reasoning}. So far, knowledge-enhanced VL literature trusts the LM-as-KB paradigm for several downstream tasks, demonstrating successful results. 

\section{VL tasks with knowledge}
\subsection{Visual Question Answering (VQA)}
In Visual Question Answering (VQA), a model receives an image \textit{I} and a textual question \textit{Q} referring to the image, and predicts a textual answer \textit{A}. The answer \textit{A} can be either selected among pre-defined candidates, viewing VQA as a \textit{classification} problem, or else be generated, thus placing VQA (free-text VQA) in the language \textit{generation} family.
Knowledge guidance can assist in addressing scenarios where standalone visual information is not adequate, as the ones presented in Figure \ref{fig:examples}.
A variety of knowledge-demanding datasets for knowledge-driven VQA (K-VQA) have been developed \cite{okvqa, vqa5, vqa1, fvqa, vqa6, vqa8, vqa21}, setting a good starting point for relevant model implementations.
Early attempts in K-VQA were heavily relying on exact matching between visual or textual concepts and KG nodes via SPARQL queries \cite{vqa0, vqa1, vqa2}, inducing errors in cases when such explicit concept mapping does not exist. Embedding representations provide a more flexible solution by retrieving similar KG facts to visual and textual concepts \cite{fvqa, vqa3, vqa5}. Nevertheless, the context-free nature of classic word embedding methods \cite{glove, word2vec} can only serve a limited amount of cases, impeding generalization to scenarios when contextualization is necessary. Transformers leveraged on the linguistic side allowed further improvements on K-VQA models \cite{vqa9, vqa19} paving the way for consequent end-to-end VL approaches.

ConceptBERT \cite{vqa7} was the first breakthrough towards a unified KVL transformer-based architecture, achieved by considering all three modalities in a joint representation, hence being able to incorporate \textbf{commonsense} knowledge in the reasoning process. \textbf{Factual} knowledge injection following the unified KVL strategy was also explored, only requiring fine-tuning to leverage the contribution of external KGs \cite{vqa18}.
Multiple knowledge senses can be fused in unified KVL architectures, such as \textbf{factual} and \textbf{visual} knowledge, in order to cross-check the validity of predicted answers \cite{vqa14}. The dynamic incorporation of external knowledge regardless its type transforms knowledge injection to a passage retrieval problem, offering advanced adaptability to relevant architectures \cite{vqa17}. Passage retrieval from Wikipedia is also followed in \cite{vqa22}, where visual cues of different granularities (global captions, image labels and scene text) are combined to retrieved facts; consequently, all linguistic information is provided to a T5 transformer model \cite{t5}, which generates the final answer \textit{A}. Similarly, \cite{vqa23} resorts to passage retrieval from external sources, but suggests joint training of the retrieval module and the T5 answer generator, contrary to prior works. Generic information obtained via passage retrieval from external knowledge sources is deemed inadequate to answer targeted visual questions. To this end, knowledge acquisition is refined by focusing on common entities present in queries, retrieved passages and images \cite{vqa24}. The combination of several KGs \cite{conceptnet, dbpedia, webchild, haspart} under a unified larger KG can facilitate knowledge retrieval, which is performed based on the linguistic similarity between contextualized questions (questions enhanced with visual captions and scene text as context) and KG facts. The enriched input is provided to a T5 transformer which finally generates the final free-text answer \textit{A} \cite{vqa25}. Multimodal passage retrieval is addressed via the proposed Multimodal Inverse Cloze Task as pre-training objective, which learns the alignment between visual and textual information from Wikipedia entries. This technique can aid K-VQA involving named entity recognition \cite{vqa26}. 

Diverging from the usage of KGs as the knowledge source at hand, first works exploiting the LM-as-KB paradigm were introduced for K-VQA. Specifically, GPT-3 \cite{gpt3} can be used to provide facts in a few-shot manner, receiving visual captions as prompts \cite{vqa15}, similar to how traditional KGs receive SPARQL queries. Performance gains can be achieved by utilizing multiple captions as prompts to a variety of pre-trained LLMs, enabling zero-shot reasoning \cite{vqa-lm1}. Using again linguistic captions as a modality mediator, \cite{vqa-lm2} leverages frozen LLMs to address zero-shot VQA. Chain of Though (CoT) prompting of LLMs is another interesting direction, which enhances explainability of the answer derivation pipeline by revealing intermediate reasoning steps \cite{vqa-lm5}.
Instead of resorting to the linguistic modality to obtain unimodal LLM prompts, other approaches opt to  fine-tune a visual encoder jointly with the LLM, so that aligned LLM-VL representations are achieved \cite{vqa-lm4}.

There are a few implementations combining explicit and implicit knowledge sources to enjoy advantages of both worlds. KRISP \cite{vqa16} leverages several external KGs \cite{conceptnet, dbpedia, haspart}, visual knowledge from Visual Genome \cite{visualgenome}, as well as implicit knowledge from BERT \cite{bert}.
REVIVE \cite{revive} deploys several visual features to retrieve knowledge from various sources, such as Wikidata and GPT-3. Visual feature guidance was proven critical towards improving the knowledge retrieval process.
Fusing both implicit and explicit knowledge in the VL reasoning process is also followed in KAT, using a refined framework that fetches information from Wikidata and GPT-3 upon which joint reasoning is performed. A transformer decoder receives the output of the reasoning module to generate the final answer  \cite{vqa-lm0}.


\subsection{Visual Commonsense Reasoning (VCR)}
Visual Commonsense Reasoning (VCR) is a task closely related to VQA. Given a challenging question \textit{Q} regarding an image \textit{I}, a VCR model is tasked to predict the answer \textit{A}, accompanied by a rationale \textit{R} \cite{vcr}. \textbf{Commonsense} knowledge can be provided from large-scale KGs, such as ConceptNet \cite{conceptnet} and ATOMIC \cite{atomic}, or dedicated datasets, such as SWAG \cite{swag}, which contains descriptions about sequences of events.

Transformer-based endeavors for knowledge-assisted VCR (K-VCR) naturally utilize BERT \cite{bert} as the backbone architecture to construct end-to-end KVL models. In KVL-BERT \cite{vcr6}, the input \textit{Q} together with candidate answers \textit{A} guide the retrieval of relevant commonsense facts  \cite{conceptnet}, resulting in a knowledge-enriched linguistic input. Then, visual features among with this enriched input are inserted in a BERT-like VL model (VL-BERT \cite{su2020vlbert}) so that the correct \textit{A} is selected. Consequently, inferring \textit{R} requires feeding VL-BERT with the predicted \textit{A}, candidate rationales \textit{R} and visual features. Aligning independent modality representations within a single multimodal embedding is proposed in \cite{vcr2}. The same work introduces extensions of VL pre-training objectives \cite{kvl-survey} to incorporate commonsense knowledge from \cite{conceptnet} as an extra modality, therefore enforcing learning KVL interrelationships. Dynamic commonsense augmentation of image-text training data is a suggested direction, accompanied by learning to reconstruct hidden visual labels based on knowledge facts retrieved from commonsense KBs \cite{vcr9}.

Implicit knowledge sources have been gaining popularity in recent K-VCR literature.
GPT-2 \cite{gpt2} has assisted dynamic reasoning over images, inferring \textbf{temporal} hypotheses regarding what might have happened before and what might happen after the depicted situation \cite{visualcomet}.
Chain of Thought (CoT) reasoning is inherently tied to VCR, as reasoning paths are highly associated with selecting rationales \textit{R}. The rise in popularity of CoT techniques for linguistic tasks is highly interconnected with the development of LLMs, which have been proven able to reveal intermediate reasoning steps \cite{cot}.
There are not yet many works in the VL direction, even though the introduction of novel appropriate datasets with grounded answer rationales highlight the prospects of such an approach \cite{vcr10}. Specifically, \cite{vcr10} tackles VCR by captioning the image, and then feed the caption together with the existing linguistic input to the LLM.
Another promising work in this direction introduces Multimodal-CoT without using language as the mediating modality, proposing a two-stage process to separately infer the answer \textit{A} and the rationale \textit{R}, while stating that a LM with less than 1B parameters is adequate for state-of-the-art performance \cite{vcr11}. 
It is expected that the rapid rise of popularity of LLMs in complex linguistic QA reasoning \cite{ComplexQA} may soon give rise to more LLM-augmented VCR approaches, addressing more aspects of reasoning.

\subsection{Image Captioning (IC)} Image Captioning (IC) is a widespread VL task, asking from a model to generate a caption \textit{c} for an image \textit{I}. Several knowledge-enhanced IC (K-IC) techniques employ recurrent neural networks (RNNs) or relevant variants such as Long Short Term Memory (LSTM) networks, while leveraging knowledge sources for \textbf{commonsense}-enhanced captioning \cite{caption1, caption2, caption3, caption5}.

The integration of external knowledge in IC using transformers, was first explored in \cite{caption10}, where \textbf{event} and \textbf{named-entity} knowledge is fused together with textual and visual data in a Transformer encoder \cite{transformer} to generate entity/event-aware captions. \textbf{Commonsense} descriptions derived from ConceptNet \cite{conceptnet} and ATOMIC \cite{atomic} are able to assist visual commonsense generation (VCG), a challenging task that requires inferring \textbf{intents} and \textbf{temporal} sequence of events \cite{caption8}. This is achieved by incorporating \textbf{commonsense} descriptions to BART, a powerful language generation model \cite{bart}. \textbf{Geographical} information guiding \textbf{factual} knowledge retrieval to assist IC was first explored in \cite{caption14}, where visual features together with the extracted facts are inserted in a Transformer encoder-decoder structure, ultimately generating the caption \textit{c}.

Apart from external knowledge considerations, IC faces additional challenges as a \textit{language generation} task: VL transformers are not well-suited for generative tasks, even though they excel in understanding tasks, where an answer has to be selected among a set of pre-defined options. XGPT tackles this challenge by adapting generative pre-training \cite{gpt2, gpt3} for VL tasks \cite{xgpt}, which is achieved by introducing novel generative pre-training objectives. The collaboration of GPT-2 \cite{gpt2} with CLIP \cite{clip} is viewed as highly promising, since both models have been trained on an abundance of web-data, thus incorporating numerous knowledge senses in an \textit{implicit} manner. ClipCap \cite{clipcap} leverages this collaboration without re-training CLIP or GPT-2; instead, a lightweight transformer-based mapping module is trained to match CLIP representations to GPT-2, which eventually generates the caption \textit{c}. The CLIP-GPT-2 combination was also followed in VC-GPT \cite{caption12}. Another lightweight improvement combining a pre-trained CLIP visual encoder and a frozen GPT-2 text decoder further boosts performance of low-resource approaches \cite{caption13}. A cross-modal filter that selects the most relevant visual information, so that captioning errors are reduced is proposed in \cite{caption11}, still respecting the frozen CLIP-GPT-2 framework.

A factor that overshadows the knowledge-enhanced IC capabilities is the lack of dedicated datasets for testing. So far, IC models are evaluated on classic datasets containing images and captions, such as COCO \cite{coco} and Flickr \cite{flickr}, which however are not challenging in terms of external knowledge required. The construction of appropriate datasets that would follow the paradigm of knowledge-demanding VQA datasets, such as OK-VQA \cite{okvqa}, K-VQA \cite{vqa5}, FVQA \cite{fvqa}, KB-VQA \cite{vqa1}, or VCR datasets \cite{vcr} will give prominence to the abilities of K-IC models. 

\subsection{Sequential Generation}
There are two tasks touching sequential generation, depending on which modality is generated at a time. Visual Storytelling (VIST) refers to generating captions $c_1, c_2, ..., c_N$ for a visual story comprised of frames $I_1, I_2,..., I_N$, as an extension of IC for sequences. The reverse task of Story Visualization (SV) involves synthesizing visual frames $I_1, I_2,..., I_N$ from textual captions $c_1, c_2, ..., c_N$. For both tasks, consistency throughout the story and relevance between the two modalities are required.

Harnessing \textbf{commonsense} knowledge for VIST was first attempted in \cite{story1}, followed by \cite{story2, story3}, which however utilize RNN-bases structures for text generation. The usage of transformer-based models is explored in \cite{story4}, where visual concepts are enriched through ConceptNet. All relevant enriched concepts are provided to BART, which ultimately outputs appropriate captions. Regarding SV, there is a different architectural line followed \cite{storygan, Maharana2021ImprovingGA, Maharana2021IntegratingVL, impartial} based on Generative Adversarial Networks (GANs) \cite{gan}. \textbf{Commonsense} and \textbf{spatial} knowledge considerations were primarily addressed in \cite{Maharana2021IntegratingVL}, demonstrating encouraging results in favor of the usage of external knowledge.
Nevertheless, some recent approaches follow transformer-based approached enhanced with \textit{implicit} knowledge. Specifically, Story-DALL-E \cite{storydalle} is able to even synthesize unseen stories in a zero-shot fashion. It leverages DALL-E \cite{dalle} as a multimodal unstructured knowledge base that provides high-quality visual synthesis from text.

Similar to IC, sequential generation tasks face the lack of appropriate datasets, upon which the external knowledge contribution would be more evident and meaningful.

\subsection{Multi-taskers}
Knowledge-free VL models have already achieved incorporating a variety of VL tasks under the same pre-training body, only requiring fine-tuning on smaller labelled text-image datasets. This way, there is no need to design and implement a separate architecture per independent task, but rather exploit visiolinguistic relationships present in large scale datasets used for pre-training, such as COCO \cite{coco}, Visual Genome \cite{visualgenome}, Conceptual Captions \cite{cc} and SBU \cite{sbu}.

\textit{Reasoning tasks}, including visual question answering, visual-text entailment and visual commonsense reasoning can be easily incorporated under the same model, due to the high similarity of the inferences these tasks have to make. Visual cues are enhanced with higher-level cognition provided by VisualCOMET \cite{visualcomet} and among with textual inputs, they are fed in a GPT-2 model that generates free-text rationales for all three tasks, therefore providing explainability of answers \cite{multi2}. The same reasoning tasks were also explored in \cite{multi3}, testing results on knowledge-demanding datasets \cite{okvqa, fvqa} as well. Knowledge embeddings representing facts from external knowledge sources \cite{conceptnet, wikidata} are aligned with textual descriptions, which are inserted together with the knowledge features in a multimodal transformer. KB-VLP \cite{multi6} is another multi-task model tackling visual question answering and visual reasoning, enhanced with \textbf{commonsense} and \textbf{logical} capabilities. Entity extraction from images and text is performed to map VL concepts to Wikidata \cite{wikidata} entries, based on knowledge graph embeddings, thus finally assisting the retrieval of the most relevant Wikidata facts.

Few-shot and zero-shot learning via LLMs in low-resource scenarios can be very practical for generative VL tasks. In \cite{vqa-lm4}, an encoder-decoder Transformer model is the most appropriate option for text generation, addressing free-form VQA and IC, while various prompts are leveraged in inference time to enforce generating a suitable textual prediction. 
PromptCap \cite{promptcap} is designed to accurately generate fine-grained and controllable captions based on GPT-3 prompting. The generated captions can be leveraged to provide context for VQA.
The Socratic model framework \cite{socratic} is a novel and promising direction, since it demonstrates that the complementary knowledge obtained during pre-training from VL and LM can be combined towards several multimodal tasks with the help of multimodal prompting.

There are some significant observations arising from the construction of multi-task VL models regarding the usage of external knowledge. For example, in \cite{multi3}, incorporation of Wikidata  failed to enhance reasoning capabilities as expected. The authors attribute this deteriorated performance to ambiguities and noise existing in the large Wikidata KB. Furthermore, comparing to task-specific implementations for VQA, IC etc, it is obvious that multi-task implementations are significantly fewer. This indicates that the incorporation of external knowledge is not as straightforward for multiple tasks as it may be for task-specific models.


 \subsection{The future of knowledge in VL}
 Recent literature around KVL research reveals several gaps that need to be covered. In our opinion, the most prominent gap is the lack of appropriate knowledge-demanding datasets for most tasks, which limits the extend to which KVL models are evaluated. Apart from VQA \cite{okvqa, vqa5, vqa1, fvqa, vqa6, vqa8, vqa21} and VCR \cite{vcr}, other VL tasks are tested on classic benchmark datasets; therefore, the contribution of knowledge in downstream performance is not as prevalent as in situations where various senses of knowledge are explicitly required.

 At the same time, we have observed that several efforts are dedicated towards constructing \textit{linguistic} benchmarks questioning reasoning capabilities in LMs, including mathematics \cite{mathqa}, symbolic reasoning \cite{cot-reasoning}, implicit reasoning based on strategies \cite{implicit-strategy}, commonsense understanding \cite{commonsenseqa},
 temporal, causal, linguistic understanding and others \cite{imitation-game}. We argue that such attempts could assist the creation of appropriate VL datasets, which would incorporate visual and linguistic challenges, so that knowledge contribution would be more concrete.

 With the surge of larger and larger models, explainability concerns are raised in the broader AI community \cite{xai-survey}, especially when black-box VL models are combined with black-box unstructured KBs. Older KVL architectures were often addressing explainability \cite{vqa2}, as an immediate result of incorporating KGs for answer prediction, since the path leading to the final answer could be retrieved. Later works, even though widely exploiting KGs, mainly focused on improving downstream performance rather than enhancing the interpretability of reasoning towards these results.
 Currently, KVL approaches have totally deviated from the pursue of explainability, especially since opaque LLMs started acting as KBs for VL models.

 Another emerging issue is prompt design for KVL architectures that exploit LLMs as their external knowledge source. Similar challenges to linguistic prompt search also arise for the VL setting (especially since many models use language as a mediator between vision and language), thus inducing some ambiguity regarding the quality of results and the reasoning process followed to return these results. Multimodal prompt design is still an unexplored but crucial field towards unlocking the full potential of LLM-VL models.
 
 Ultimately, we view that the future of KVL research is highly interconnected with the current trends in LLMs, both in terms of designing appropriate knowledge-demanding benchmarks for KVL tasks, as well as answering the KG vs LLM ongoing dilemma, analyzed in Section \ref{sec:kb-or-llm}.

\section{Knowledge Graphs or Large Language Models?}
\label{sec:kb-or-llm}
Throughout our analysis we recognize some potential trends towards selecting the type of knowledge source to assist VL models towards hybrid approaches. Even though KGs clearly dominate previous VL implementations, we can assume some focus shifting towards LMs-as-KB, due to the rapid development of L(L)Ms and generally their ever increasing popularity in recent NLP literature. For example, the advent of ChatGPT\footnote{\href{https://openai.com/blog/chatgpt/}{ChatGPT}} created an unprecedented hype, opening several discussions regarding the future of AI as a whole. The almost human-level capabilities of ChatGPT and relevant implementations have created an abundance of opportunities in NLP literature, while sparking a lot of unavoidable criticism.
Since VL learning tends to follow the advancements in NLP, as it happened with KG-based knowledge boosting \cite{luke, erica, ebert, ernie-nli}, it is highly likely that more LLM-based VL implementations will soon emerge. 

However, this trend comes at a cost: LLMs have already scaled to billions and even trillions \cite{glam} of parameters, a procedure that requires massive training.
Concerns regarding the cost of pre-training language models has been raised even before this tremendous parameter scaling \cite{pretraining-cost}; apart from computational budget, issues such as fair access and environmental impact of relevant implementations should question the reliance and preference of the research community towards them. 

In the meanwhile, the completely opaque learning and decision-making process may be more harmful than beneficial; even though known biases, errors and inaccuracies\footnote[2]{\href{https://openai.com/blog/instruction-following/\#birds-not-real}{Birds are not real}}
are reported in LLM publications, their actual usage raises doubts regarding the quality and the trustworthiness of the knowledge provided to the final task. 
 Early on the introduction of LLMs, such as GPT-3 \cite{gpt3}, papers exposing failure cases regarding mathematical reasoning, logic and ethical requests \cite{gpt3-kraksimo} gained lots of attention.
Probing how LLMs can be fooled, sheds some light to their reasoning process and how they are being confused by misleading inputs \cite{factcheck, distracted}, which can be a promising starting point towards defeating logical brittleness. 
Relevant endeavors uncover LLM deductive reasoning capabilities \cite{deductive}, proving that exhaustive memorization compensates for LLM inability to learn to reason.
In total, reasoning capabilities of LLMs pose several open questions \cite{reasoning}, such as whether heuristics conceal reasoning incapability, or whether reasoning steps can be trustworthy, given that inconsistencies and false rationales are sometimes provided with certainty.
More refined challenges reveal that LLMs' world-knowledge suffers from robustness issues as well, since they confuse likely and unlikely situations, even though they are capable of recognizing impossible events \cite{unlikely}.

Even if aforementioned concerns are somehow addressed in consequent versions of relevant  LLMs, the search for the optimal fact retrieval process via prompts still remains an open challenge \cite{prompt-survey}. Moreover, papers promoting certain prompts to LLMs avoid describing the process behind discovering such optimal prompts, and merely provide some experimental comparison between similar prompt phrases \cite{cot}. Therefore, if the golden prompt \cite{cot} \textit{let's think step by step} has been defined via extended experimentation, there is no guarantee regarding its optimality and reliability; on the other hand, if there is a certain methodology behind, it seems that it has not been fully unlocked (or at least released to the public). Low-performance instructive prompts targeting CoT reasoning are semantically relevant to the golden prompt, thus raising doubts regarding the consistency of LLM-occurring answers and their sensitivity to -slight or more intense- input variations.

Overall, the aforementioned shortcomings are expected to be transferred to VL implementations if LLMs eventually replace KGs to some extend, thus raising a crucial question: is it worth it to quickly adapt to the trend or better wait? 

\section{Conclusion}
In this survey, we analyzed the collaboration of external knowledge with VL learning. Existing models and datasets drive several challenges in this upcoming field, since the full potential of knowledge-enhanced approaches has not been yet unlocked. Knowledge graphs and large language models can both serve as knowledge bases for VL, posing different advantages and disadvantages. To this end, the current paper devotes an extended discussion over the KB vs LLM dilemma for VL, highlighting significant open issues tied to the current state of the LM-as-KB paradigm. All in all, we hope that our work can introduce researchers to the knowledge-enhanced VL exploration, while denoting challenges of the knowledge adoption process.

\begin{acknowledgments}
The research work was supported by the Hellenic Foundation for Research and Innovation (HFRI) under the 3rd Call for HFRI PhD Fellowships (Fellowship Number 5537).  
\end{acknowledgments}

\bibliography{sample-ceur}

\begin{thebibliography}{150}
\expandafter\ifx\csname natexlab\endcsname\relax\def\natexlab#1{#1}\fi
\providecommand{\url}[1]{\texttt{#1}}
\providecommand{\href}[2]{#2}
\providecommand{\path}[1]{#1}
\providecommand{\DOIprefix}{doi:}
\providecommand{\ArXivprefix}{arXiv:}
\providecommand{\URLprefix}{URL: }
\providecommand{\Pubmedprefix}{pmid:}
\providecommand{\doi}[1]{\href{http://dx.doi.org/#1}{\path{#1}}}
\providecommand{\Pubmed}[1]{\href{pmid:#1}{\path{#1}}}
\providecommand{\bibinfo}[2]{#2}
\ifx\xfnm\relax \def\xfnm[#1]{\unskip,\space#1}\fi
\bibitem[{Vaswani et~al.(2017)Vaswani, Shazeer, Parmar, Uszkoreit, Jones,
  Gomez, Kaiser, and Polosukhin}]{transformer}
\bibinfo{author}{A.~Vaswani}, \bibinfo{author}{N.~Shazeer},
  \bibinfo{author}{N.~Parmar}, \bibinfo{author}{J.~Uszkoreit},
  \bibinfo{author}{L.~Jones}, \bibinfo{author}{A.~N. Gomez},
  \bibinfo{author}{L.~u. Kaiser}, \bibinfo{author}{I.~Polosukhin},
\newblock \bibinfo{title}{Attention is all you need},
\newblock in: \bibinfo{editor}{I.~Guyon}, \bibinfo{editor}{U.~V. Luxburg},
  \bibinfo{editor}{S.~Bengio}, \bibinfo{editor}{H.~Wallach},
  \bibinfo{editor}{R.~Fergus}, \bibinfo{editor}{S.~Vishwanathan},
  \bibinfo{editor}{R.~Garnett} (Eds.), \bibinfo{booktitle}{Advances in Neural
  Information Processing Systems}, volume~\bibinfo{volume}{30},
  \bibinfo{publisher}{Curran Associates, Inc.}, \bibinfo{year}{2017}.
  \URLprefix
  \url{https://proceedings.neurips.cc/paper/2017/file/3f5ee243547dee91fbd053c1c4a845aa-Paper.pdf}.
\bibitem[{Agrawal et~al.(2016)Agrawal, Lu, Antol, Mitchell, Zitnick, Batra, and
  Parikh}]{vqa}
\bibinfo{author}{A.~Agrawal}, \bibinfo{author}{J.~Lu},
  \bibinfo{author}{S.~Antol}, \bibinfo{author}{M.~Mitchell},
  \bibinfo{author}{C.~L. Zitnick}, \bibinfo{author}{D.~Batra},
  \bibinfo{author}{D.~Parikh}, \bibinfo{title}{Vqa: Visual question answering},
  \bibinfo{year}{2016}. \href{http://arxiv.org/abs/1505.00468}{{\tt
  arXiv:1505.00468}}.
\bibitem[{He et~al.(2021)He, Wang, Miao, and Sun}]{visual_reasoning}
\bibinfo{author}{F.~He}, \bibinfo{author}{Y.~Wang}, \bibinfo{author}{X.~Miao},
  \bibinfo{author}{X.~Sun},
\newblock \bibinfo{title}{Interpretable visual reasoning: A survey},
\newblock \bibinfo{journal}{Image and Vision Computing} \bibinfo{volume}{112}
  (\bibinfo{year}{2021}) \bibinfo{pages}{104194}. \URLprefix
  \url{https://www.sciencedirect.com/science/article/pii/S0262885621000998}.
  \DOIprefix\doi{https://doi.org/10.1016/j.imavis.2021.104194}.
\bibitem[{Zellers et~al.(2019)Zellers, Bisk, Farhadi, and Choi}]{vcr}
\bibinfo{author}{R.~Zellers}, \bibinfo{author}{Y.~Bisk},
  \bibinfo{author}{A.~Farhadi}, \bibinfo{author}{Y.~Choi}, \bibinfo{title}{From
  recognition to cognition: Visual commonsense reasoning},
  \bibinfo{year}{2019}. \href{http://arxiv.org/abs/1811.10830}{{\tt
  arXiv:1811.10830}}.
\bibitem[{Xie et~al.(2018)Xie, Lai, Doran, and Kadav}]{entailment}
\bibinfo{author}{N.~Xie}, \bibinfo{author}{F.~Lai}, \bibinfo{author}{D.~Doran},
  \bibinfo{author}{A.~Kadav},
\newblock \bibinfo{title}{Visual entailment task for visually-grounded language
  learning},
\newblock \bibinfo{journal}{arXiv preprint arXiv:1811.10582}
  (\bibinfo{year}{2018}).
\bibitem[{Stefanini et~al.(2021)Stefanini, Cornia, Baraldi, Cascianelli,
  Fiameni, and Cucchiara}]{captioning}
\bibinfo{author}{M.~Stefanini}, \bibinfo{author}{M.~Cornia},
  \bibinfo{author}{L.~Baraldi}, \bibinfo{author}{S.~Cascianelli},
  \bibinfo{author}{G.~Fiameni}, \bibinfo{author}{R.~Cucchiara},
  \bibinfo{title}{From show to tell: A survey on deep learning-based image
  captioning}, \bibinfo{year}{2021}.
  \href{http://arxiv.org/abs/2107.06912}{{\tt arXiv:2107.06912}}.
\bibitem[{Dubey(2021)}]{retrieval}
\bibinfo{author}{S.~R. Dubey},
\newblock \bibinfo{title}{A decade survey of content based image retrieval
  using deep learning},
\newblock \bibinfo{journal}{IEEE Transactions on Circuits and Systems for Video
  Technology}  (\bibinfo{year}{2021}) \bibinfo{pages}{1–1}. \URLprefix
  \url{http://dx.doi.org/10.1109/TCSVT.2021.3080920}.
  \DOIprefix\doi{10.1109/tcsvt.2021.3080920}.
\bibitem[{Anderson et~al.(2018)Anderson, Wu, Teney, Bruce, Johnson,
  Sünderhauf, Reid, Gould, and van~den Hengel}]{navigation}
\bibinfo{author}{P.~Anderson}, \bibinfo{author}{Q.~Wu},
  \bibinfo{author}{D.~Teney}, \bibinfo{author}{J.~Bruce},
  \bibinfo{author}{M.~Johnson}, \bibinfo{author}{N.~Sünderhauf},
  \bibinfo{author}{I.~Reid}, \bibinfo{author}{S.~Gould},
  \bibinfo{author}{A.~van~den Hengel}, \bibinfo{title}{Vision-and-language
  navigation: Interpreting visually-grounded navigation instructions in real
  environments}, \bibinfo{year}{2018}.
  \href{http://arxiv.org/abs/1711.07280}{{\tt arXiv:1711.07280}}.
\bibitem[{El-Nouby et~al.(2019)El-Nouby, Sharma, Schulz, Hjelm, Asri, Kahou,
  Bengio, and Taylor}]{dialog-gen}
\bibinfo{author}{A.~El-Nouby}, \bibinfo{author}{S.~Sharma},
  \bibinfo{author}{H.~Schulz}, \bibinfo{author}{D.~Hjelm},
  \bibinfo{author}{L.~E. Asri}, \bibinfo{author}{S.~E. Kahou},
  \bibinfo{author}{Y.~Bengio}, \bibinfo{author}{G.~W. Taylor},
  \bibinfo{title}{Tell, draw, and repeat: Generating and modifying images based
  on continual linguistic instruction}, \bibinfo{year}{2019}.
  \href{http://arxiv.org/abs/1811.09845}{{\tt arXiv:1811.09845}}.
\bibitem[{Lu et~al.(2019)Lu, Batra, Parikh, and Lee}]{lu2019vilbert}
\bibinfo{author}{J.~Lu}, \bibinfo{author}{D.~Batra},
  \bibinfo{author}{D.~Parikh}, \bibinfo{author}{S.~Lee},
  \bibinfo{title}{Vilbert: Pretraining task-agnostic visiolinguistic
  representations for vision-and-language tasks}, \bibinfo{year}{2019}.
  \href{http://arxiv.org/abs/1908.02265}{{\tt arXiv:1908.02265}}.
\bibitem[{Li et~al.(2020)Li, Yin, Li, Zhang, Hu, Zhang, Wang, Hu, Dong, Wei,
  Choi, and Gao}]{li2020oscar}
\bibinfo{author}{X.~Li}, \bibinfo{author}{X.~Yin}, \bibinfo{author}{C.~Li},
  \bibinfo{author}{P.~Zhang}, \bibinfo{author}{X.~Hu},
  \bibinfo{author}{L.~Zhang}, \bibinfo{author}{L.~Wang},
  \bibinfo{author}{H.~Hu}, \bibinfo{author}{L.~Dong}, \bibinfo{author}{F.~Wei},
  \bibinfo{author}{Y.~Choi}, \bibinfo{author}{J.~Gao}, \bibinfo{title}{Oscar:
  Object-semantics aligned pre-training for vision-language tasks},
  \bibinfo{year}{2020}. \href{http://arxiv.org/abs/2004.06165}{{\tt
  arXiv:2004.06165}}.
\bibitem[{Singh et~al.(2021)Singh, Hu, Goswami, Couairon, Galuba, Rohrbach, and
  Kiela}]{flava}
\bibinfo{author}{A.~Singh}, \bibinfo{author}{R.~Hu},
  \bibinfo{author}{V.~Goswami}, \bibinfo{author}{G.~Couairon},
  \bibinfo{author}{W.~Galuba}, \bibinfo{author}{M.~Rohrbach},
  \bibinfo{author}{D.~Kiela},
\newblock \bibinfo{title}{Flava: A foundational language and vision alignment
  model}  (\bibinfo{year}{2021}).
\bibitem[{Kim et~al.(2021)Kim, Son, and Kim}]{kim2021vilt}
\bibinfo{author}{W.~Kim}, \bibinfo{author}{B.~Son}, \bibinfo{author}{I.~Kim},
  \bibinfo{title}{Vilt: Vision-and-language transformer without convolution or
  region supervision}, \bibinfo{year}{2021}.
  \href{http://arxiv.org/abs/2102.03334}{{\tt arXiv:2102.03334}}.
\bibitem[{Wang et~al.(2021)Wang, Yu, Yu, Dai, Tsvetkov, and Cao}]{simvlm}
\bibinfo{author}{Z.~Wang}, \bibinfo{author}{J.~Yu}, \bibinfo{author}{A.~W. Yu},
  \bibinfo{author}{Z.~Dai}, \bibinfo{author}{Y.~Tsvetkov},
  \bibinfo{author}{Y.~Cao}, \bibinfo{title}{Simvlm: Simple visual language
  model pretraining with weak supervision}, \bibinfo{year}{2021}. \URLprefix
  \url{https://arxiv.org/abs/2108.10904}.
  \DOIprefix\doi{10.48550/ARXIV.2108.10904}.
\bibitem[{Radford et~al.(2021)Radford, Kim, Hallacy, Ramesh, Goh, Agarwal,
  Sastry, Askell, Mishkin, Clark, Krueger, and Sutskever}]{clip}
\bibinfo{author}{A.~Radford}, \bibinfo{author}{J.~Kim},
  \bibinfo{author}{C.~Hallacy}, \bibinfo{author}{A.~Ramesh},
  \bibinfo{author}{G.~Goh}, \bibinfo{author}{S.~Agarwal},
  \bibinfo{author}{G.~Sastry}, \bibinfo{author}{A.~Askell},
  \bibinfo{author}{P.~Mishkin}, \bibinfo{author}{J.~Clark},
  \bibinfo{author}{G.~Krueger}, \bibinfo{author}{I.~Sutskever},
  \bibinfo{title}{Learning transferable visual models from natural language
  supervision}, \bibinfo{year}{2021}.
\bibitem[{Huang et~al.(2021)Huang, Zeng, Huang, Liu, Fu, and Fu}]{soho}
\bibinfo{author}{Z.~Huang}, \bibinfo{author}{Z.~Zeng},
  \bibinfo{author}{Y.~Huang}, \bibinfo{author}{B.~Liu},
  \bibinfo{author}{D.~Fu}, \bibinfo{author}{J.~Fu}, \bibinfo{title}{Seeing out
  of the box: End-to-end pre-training for vision-language representation
  learning}, \bibinfo{year}{2021}. \URLprefix
  \url{https://arxiv.org/abs/2104.03135}.
  \DOIprefix\doi{10.48550/ARXIV.2104.03135}.
\bibitem[{Yuan et~al.(2021)Yuan, Chen, Chen, Codella, Dai, Gao, Hu, Huang, Li,
  Li, Liu, Liu, Liu, Lu, Shi, Wang, Wang, Xiao, Xiao, Yang, Zeng, Zhou, and
  Zhang}]{florence}
\bibinfo{author}{L.~Yuan}, \bibinfo{author}{D.~Chen}, \bibinfo{author}{Y.-L.
  Chen}, \bibinfo{author}{N.~Codella}, \bibinfo{author}{X.~Dai},
  \bibinfo{author}{J.~Gao}, \bibinfo{author}{H.~Hu},
  \bibinfo{author}{X.~Huang}, \bibinfo{author}{B.~Li}, \bibinfo{author}{C.~Li},
  \bibinfo{author}{C.~Liu}, \bibinfo{author}{M.~Liu}, \bibinfo{author}{Z.~Liu},
  \bibinfo{author}{Y.~Lu}, \bibinfo{author}{Y.~Shi}, \bibinfo{author}{L.~Wang},
  \bibinfo{author}{J.~Wang}, \bibinfo{author}{B.~Xiao},
  \bibinfo{author}{Z.~Xiao}, \bibinfo{author}{J.~Yang},
  \bibinfo{author}{M.~Zeng}, \bibinfo{author}{L.~Zhou},
  \bibinfo{author}{P.~Zhang}, \bibinfo{title}{Florence: A new foundation model
  for computer vision}, \bibinfo{year}{2021}. \URLprefix
  \url{https://arxiv.org/abs/2111.11432}.
  \DOIprefix\doi{10.48550/ARXIV.2111.11432}.
\bibitem[{Jia et~al.(2021)Jia, Yang, Xia, Chen, Parekh, Pham, Le, Sung, Li, and
  Duerig}]{align}
\bibinfo{author}{C.~Jia}, \bibinfo{author}{Y.~Yang}, \bibinfo{author}{Y.~Xia},
  \bibinfo{author}{Y.-T. Chen}, \bibinfo{author}{Z.~Parekh},
  \bibinfo{author}{H.~Pham}, \bibinfo{author}{Q.~V. Le},
  \bibinfo{author}{Y.~Sung}, \bibinfo{author}{Z.~Li},
  \bibinfo{author}{T.~Duerig},
\newblock \bibinfo{title}{Scaling up visual and vision-language representation
  learning with noisy text supervision}  (\bibinfo{year}{2021}). \URLprefix
  \url{https://arxiv.org/abs/2102.05918}.
  \DOIprefix\doi{10.48550/ARXIV.2102.05918}.
\bibitem[{Marino et~al.(2019)Marino, Rastegari, Farhadi, and Mottaghi}]{okvqa}
\bibinfo{author}{K.~Marino}, \bibinfo{author}{M.~Rastegari},
  \bibinfo{author}{A.~Farhadi}, \bibinfo{author}{R.~Mottaghi},
\newblock \bibinfo{title}{Ok-vqa: A visual question answering benchmark
  requiring external knowledge},
\newblock \bibinfo{journal}{2019 IEEE/CVF Conference on Computer Vision and
  Pattern Recognition (CVPR)}  (\bibinfo{year}{2019})
  \bibinfo{pages}{3190--3199}.
\bibitem[{Shah et~al.(2019)Shah, Mishra, Yadati, and Talukdar}]{vqa5}
\bibinfo{author}{S.~Shah}, \bibinfo{author}{A.~Mishra},
  \bibinfo{author}{N.~Yadati}, \bibinfo{author}{P.~P. Talukdar},
\newblock \bibinfo{title}{Kvqa: Knowledge-aware visual question answering},
\newblock in: \bibinfo{booktitle}{AAAI}, \bibinfo{year}{2019}.
\bibitem[{Garcia-Olano et~al.(2021)Garcia-Olano, Onoe, and Ghosh}]{vqa18}
\bibinfo{author}{D.~Garcia-Olano}, \bibinfo{author}{Y.~Onoe},
  \bibinfo{author}{J.~Ghosh},
\newblock \bibinfo{title}{Improving and diagnosing knowledge-based visual
  question answering via entity enhanced knowledge injection},
\newblock \bibinfo{year}{2021}.
\bibitem[{Ji et~al.(2021)Ji, Pan, Cambria, Marttinen, and Yu}]{kg1}
\bibinfo{author}{S.~Ji}, \bibinfo{author}{S.~Pan},
  \bibinfo{author}{E.~Cambria}, \bibinfo{author}{P.~Marttinen},
  \bibinfo{author}{P.~S. Yu},
\newblock \bibinfo{title}{A survey on knowledge graphs: Representation,
  acquisition and applications},
\newblock \bibinfo{journal}{IEEE transactions on neural networks and learning
  systems} \bibinfo{volume}{PP} (\bibinfo{year}{2021}).
\bibitem[{Fellbaum(1998)}]{wordnet}
\bibinfo{author}{C.~Fellbaum},
\newblock \bibinfo{title}{Wordnet: An electronic lexical database}
  (\bibinfo{year}{1998}).
\bibitem[{Speer et~al.(2017)Speer, Chin, and Havasi}]{conceptnet}
\bibinfo{author}{R.~Speer}, \bibinfo{author}{J.~Chin},
  \bibinfo{author}{C.~Havasi},
\newblock \bibinfo{title}{Conceptnet 5.5: An open multilingual graph of general
  knowledge},
\newblock in: \bibinfo{booktitle}{AAAI}, \bibinfo{year}{2017}.
\bibitem[{Vrande\u{c}i\`{c} and Kr\"{o}tzsch(2014)}]{wikidata}
\bibinfo{author}{D.~Vrande\u{c}i\`{c}}, \bibinfo{author}{M.~Kr\"{o}tzsch},
\newblock \bibinfo{title}{Wikidata: a free collaborative knowledgebase},
\newblock \bibinfo{journal}{Commun. ACM} \bibinfo{volume}{57}
  (\bibinfo{year}{2014}) \bibinfo{pages}{78--85}.
\bibitem[{Auer et~al.(2007)Auer, Bizer, Kobilarov, Lehmann, Cyganiak, and
  Ives}]{dbpedia}
\bibinfo{author}{S.~Auer}, \bibinfo{author}{C.~Bizer},
  \bibinfo{author}{G.~Kobilarov}, \bibinfo{author}{J.~Lehmann},
  \bibinfo{author}{R.~Cyganiak}, \bibinfo{author}{Z.~G. Ives},
\newblock \bibinfo{title}{Dbpedia: A nucleus for a web of open data},
\newblock in: \bibinfo{booktitle}{ISWC/ASWC}, \bibinfo{year}{2007}.
\bibitem[{Devlin et~al.(2019)Devlin, Chang, Lee, and Toutanova}]{bert}
\bibinfo{author}{J.~Devlin}, \bibinfo{author}{M.-W. Chang},
  \bibinfo{author}{K.~Lee}, \bibinfo{author}{K.~Toutanova},
\newblock \bibinfo{title}{Bert: Pre-training of deep bidirectional transformers
  for language understanding},
\newblock \bibinfo{journal}{ArXiv} \bibinfo{volume}{abs/1810.04805}
  (\bibinfo{year}{2019}).
\bibitem[{Petroni et~al.(2019)Petroni, Rocktäschel, Lewis, Bakhtin, Wu,
  Miller, and Riedel}]{lmkb1}
\bibinfo{author}{F.~Petroni}, \bibinfo{author}{T.~Rocktäschel},
  \bibinfo{author}{P.~Lewis}, \bibinfo{author}{A.~Bakhtin},
  \bibinfo{author}{Y.~Wu}, \bibinfo{author}{A.~H. Miller},
  \bibinfo{author}{S.~Riedel}, \bibinfo{title}{Language models as knowledge
  bases?}, \bibinfo{year}{2019}. \URLprefix
  \url{https://arxiv.org/abs/1909.01066}.
  \DOIprefix\doi{10.48550/ARXIV.1909.01066}.
\bibitem[{Wang et~al.(2020)Wang, Liu, and Song}]{lmkb3}
\bibinfo{author}{C.~Wang}, \bibinfo{author}{X.~Liu}, \bibinfo{author}{D.~Song},
  \bibinfo{title}{Language models are open knowledge graphs},
  \bibinfo{year}{2020}. \URLprefix \url{https://arxiv.org/abs/2010.11967}.
  \DOIprefix\doi{10.48550/ARXIV.2010.11967}.
\bibitem[{AlKhamissi et~al.(2022)AlKhamissi, Li, Celikyilmaz, Diab, and
  Ghazvininejad}]{lmkb2}
\bibinfo{author}{B.~AlKhamissi}, \bibinfo{author}{M.~Li},
  \bibinfo{author}{A.~Celikyilmaz}, \bibinfo{author}{M.~Diab},
  \bibinfo{author}{M.~Ghazvininejad}, \bibinfo{title}{A review on language
  models as knowledge bases}, \bibinfo{year}{2022}. \URLprefix
  \url{https://arxiv.org/abs/2204.06031}.
  \DOIprefix\doi{10.48550/ARXIV.2204.06031}.
\bibitem[{Brown et~al.(2020)Brown, Mann, Ryder, Subbiah, Kaplan, Dhariwal,
  Neelakantan, Shyam, Sastry, Askell, Agarwal, Herbert-Voss, Krueger, Henighan,
  Child, Ramesh, Ziegler, Wu, Winter, Hesse, Chen, Sigler, Litwin, Gray, Chess,
  Clark, Berner, McCandlish, Radford, Sutskever, and Amodei}]{gpt3}
\bibinfo{author}{T.~Brown}, \bibinfo{author}{B.~Mann},
  \bibinfo{author}{N.~Ryder}, \bibinfo{author}{M.~Subbiah},
  \bibinfo{author}{J.~D. Kaplan}, \bibinfo{author}{P.~Dhariwal},
  \bibinfo{author}{A.~Neelakantan}, \bibinfo{author}{P.~Shyam},
  \bibinfo{author}{G.~Sastry}, \bibinfo{author}{A.~Askell},
  \bibinfo{author}{S.~Agarwal}, \bibinfo{author}{A.~Herbert-Voss},
  \bibinfo{author}{G.~Krueger}, \bibinfo{author}{T.~Henighan},
  \bibinfo{author}{R.~Child}, \bibinfo{author}{A.~Ramesh},
  \bibinfo{author}{D.~Ziegler}, \bibinfo{author}{J.~Wu},
  \bibinfo{author}{C.~Winter}, \bibinfo{author}{C.~Hesse},
  \bibinfo{author}{M.~Chen}, \bibinfo{author}{E.~Sigler},
  \bibinfo{author}{M.~Litwin}, \bibinfo{author}{S.~Gray},
  \bibinfo{author}{B.~Chess}, \bibinfo{author}{J.~Clark},
  \bibinfo{author}{C.~Berner}, \bibinfo{author}{S.~McCandlish},
  \bibinfo{author}{A.~Radford}, \bibinfo{author}{I.~Sutskever},
  \bibinfo{author}{D.~Amodei},
\newblock \bibinfo{title}{Language models are few-shot learners},
\newblock in: \bibinfo{editor}{H.~Larochelle}, \bibinfo{editor}{M.~Ranzato},
  \bibinfo{editor}{R.~Hadsell}, \bibinfo{editor}{M.~F. Balcan},
  \bibinfo{editor}{H.~Lin} (Eds.), \bibinfo{booktitle}{Advances in Neural
  Information Processing Systems}, volume~\bibinfo{volume}{33},
  \bibinfo{publisher}{Curran Associates, Inc.}, \bibinfo{year}{2020}, pp.
  \bibinfo{pages}{1877--1901}. \URLprefix
  \url{https://proceedings.neurips.cc/paper/2020/file/1457c0d6bfcb4967418bfb8ac142f64a-Paper.pdf}.
\bibitem[{Hoffmann et~al.(2022)Hoffmann, Borgeaud, Mensch, Buchatskaya, Cai,
  Rutherford, Casas, Hendricks, Welbl, Clark, Hennigan, Noland, Millican,
  Driessche, Damoc, Guy, Osindero, Simonyan, Elsen, Rae, Vinyals, and
  Sifre}]{chinchilla}
\bibinfo{author}{J.~Hoffmann}, \bibinfo{author}{S.~Borgeaud},
  \bibinfo{author}{A.~Mensch}, \bibinfo{author}{E.~Buchatskaya},
  \bibinfo{author}{T.~Cai}, \bibinfo{author}{E.~Rutherford},
  \bibinfo{author}{D.~d.~L. Casas}, \bibinfo{author}{L.~A. Hendricks},
  \bibinfo{author}{J.~Welbl}, \bibinfo{author}{A.~Clark},
  \bibinfo{author}{T.~Hennigan}, \bibinfo{author}{E.~Noland},
  \bibinfo{author}{K.~Millican}, \bibinfo{author}{G.~v.~d. Driessche},
  \bibinfo{author}{B.~Damoc}, \bibinfo{author}{A.~Guy},
  \bibinfo{author}{S.~Osindero}, \bibinfo{author}{K.~Simonyan},
  \bibinfo{author}{E.~Elsen}, \bibinfo{author}{J.~W. Rae},
  \bibinfo{author}{O.~Vinyals}, \bibinfo{author}{L.~Sifre},
  \bibinfo{title}{Training compute-optimal large language models},
  \bibinfo{year}{2022}. \URLprefix \url{https://arxiv.org/abs/2203.15556}.
  \DOIprefix\doi{10.48550/ARXIV.2203.15556}.
\bibitem[{Chowdhery et~al.(2022)Chowdhery, Narang, Devlin, Bosma, Mishra,
  Roberts, Barham, Chung, Sutton, Gehrmann, Schuh, Shi, Tsvyashchenko, Maynez,
  Rao, Barnes, Tay, Shazeer, Prabhakaran, Reif, Du, Hutchinson, Pope, Bradbury,
  Austin, Isard, Gur-Ari, Yin, Duke, Levskaya, Ghemawat, Dev, Michalewski,
  Garcia, Misra, Robinson, Fedus, Zhou, Ippolito, Luan, Lim, Zoph, Spiridonov,
  Sepassi, Dohan, Agrawal, Omernick, Dai, Pillai, Pellat, Lewkowycz, Moreira,
  Child, Polozov, Lee, Zhou, Wang, Saeta, Diaz, Firat, Catasta, Wei,
  Meier-Hellstern, Eck, Dean, Petrov, and Fiedel}]{palm}
\bibinfo{author}{A.~Chowdhery}, \bibinfo{author}{S.~Narang},
  \bibinfo{author}{J.~Devlin}, \bibinfo{author}{M.~Bosma},
  \bibinfo{author}{G.~Mishra}, \bibinfo{author}{A.~Roberts},
  \bibinfo{author}{P.~Barham}, \bibinfo{author}{H.~W. Chung},
  \bibinfo{author}{C.~Sutton}, \bibinfo{author}{S.~Gehrmann},
  \bibinfo{author}{P.~Schuh}, \bibinfo{author}{K.~Shi},
  \bibinfo{author}{S.~Tsvyashchenko}, \bibinfo{author}{J.~Maynez},
  \bibinfo{author}{A.~Rao}, \bibinfo{author}{P.~Barnes},
  \bibinfo{author}{Y.~Tay}, \bibinfo{author}{N.~Shazeer},
  \bibinfo{author}{V.~Prabhakaran}, \bibinfo{author}{E.~Reif},
  \bibinfo{author}{N.~Du}, \bibinfo{author}{B.~Hutchinson},
  \bibinfo{author}{R.~Pope}, \bibinfo{author}{J.~Bradbury},
  \bibinfo{author}{J.~Austin}, \bibinfo{author}{M.~Isard},
  \bibinfo{author}{G.~Gur-Ari}, \bibinfo{author}{P.~Yin},
  \bibinfo{author}{T.~Duke}, \bibinfo{author}{A.~Levskaya},
  \bibinfo{author}{S.~Ghemawat}, \bibinfo{author}{S.~Dev},
  \bibinfo{author}{H.~Michalewski}, \bibinfo{author}{X.~Garcia},
  \bibinfo{author}{V.~Misra}, \bibinfo{author}{K.~Robinson},
  \bibinfo{author}{L.~Fedus}, \bibinfo{author}{D.~Zhou},
  \bibinfo{author}{D.~Ippolito}, \bibinfo{author}{D.~Luan},
  \bibinfo{author}{H.~Lim}, \bibinfo{author}{B.~Zoph},
  \bibinfo{author}{A.~Spiridonov}, \bibinfo{author}{R.~Sepassi},
  \bibinfo{author}{D.~Dohan}, \bibinfo{author}{S.~Agrawal},
  \bibinfo{author}{M.~Omernick}, \bibinfo{author}{A.~M. Dai},
  \bibinfo{author}{T.~S. Pillai}, \bibinfo{author}{M.~Pellat},
  \bibinfo{author}{A.~Lewkowycz}, \bibinfo{author}{E.~Moreira},
  \bibinfo{author}{R.~Child}, \bibinfo{author}{O.~Polozov},
  \bibinfo{author}{K.~Lee}, \bibinfo{author}{Z.~Zhou},
  \bibinfo{author}{X.~Wang}, \bibinfo{author}{B.~Saeta},
  \bibinfo{author}{M.~Diaz}, \bibinfo{author}{O.~Firat},
  \bibinfo{author}{M.~Catasta}, \bibinfo{author}{J.~Wei},
  \bibinfo{author}{K.~Meier-Hellstern}, \bibinfo{author}{D.~Eck},
  \bibinfo{author}{J.~Dean}, \bibinfo{author}{S.~Petrov},
  \bibinfo{author}{N.~Fiedel}, \bibinfo{title}{Palm: Scaling language modeling
  with pathways}, \bibinfo{year}{2022}. \URLprefix
  \url{https://arxiv.org/abs/2204.02311}.
  \DOIprefix\doi{10.48550/ARXIV.2204.02311}.
\bibitem[{Du et~al.(2021)Du, Huang, Dai, Tong, Lepikhin, Xu, Krikun, Zhou, Yu,
  Firat, Zoph, Fedus, Bosma, Zhou, Wang, Wang, Webster, Pellat, Robinson,
  Meier-Hellstern, Duke, Dixon, Zhang, Le, Wu, Chen, and Cui}]{glam}
\bibinfo{author}{N.~Du}, \bibinfo{author}{Y.~Huang}, \bibinfo{author}{A.~M.
  Dai}, \bibinfo{author}{S.~Tong}, \bibinfo{author}{D.~Lepikhin},
  \bibinfo{author}{Y.~Xu}, \bibinfo{author}{M.~Krikun},
  \bibinfo{author}{Y.~Zhou}, \bibinfo{author}{A.~W. Yu},
  \bibinfo{author}{O.~Firat}, \bibinfo{author}{B.~Zoph},
  \bibinfo{author}{L.~Fedus}, \bibinfo{author}{M.~Bosma},
  \bibinfo{author}{Z.~Zhou}, \bibinfo{author}{T.~Wang}, \bibinfo{author}{Y.~E.
  Wang}, \bibinfo{author}{K.~Webster}, \bibinfo{author}{M.~Pellat},
  \bibinfo{author}{K.~Robinson}, \bibinfo{author}{K.~Meier-Hellstern},
  \bibinfo{author}{T.~Duke}, \bibinfo{author}{L.~Dixon},
  \bibinfo{author}{K.~Zhang}, \bibinfo{author}{Q.~V. Le},
  \bibinfo{author}{Y.~Wu}, \bibinfo{author}{Z.~Chen}, \bibinfo{author}{C.~Cui},
  \bibinfo{title}{Glam: Efficient scaling of language models with
  mixture-of-experts}, \bibinfo{year}{2021}. \URLprefix
  \url{https://arxiv.org/abs/2112.06905}.
  \DOIprefix\doi{10.48550/ARXIV.2112.06905}.
\bibitem[{Borgeaud et~al.(2021)Borgeaud, Mensch, Hoffmann, Cai, Rutherford,
  Millican, Driessche, Lespiau, Damoc, Clark, Casas, Guy, Menick, Ring,
  Hennigan, Huang, Maggiore, Jones, Cassirer, Brock, Paganini, Irving, Vinyals,
  Osindero, Simonyan, Rae, Elsen, and Sifre}]{gopher}
\bibinfo{author}{S.~Borgeaud}, \bibinfo{author}{A.~Mensch},
  \bibinfo{author}{J.~Hoffmann}, \bibinfo{author}{T.~Cai},
  \bibinfo{author}{E.~Rutherford}, \bibinfo{author}{K.~Millican},
  \bibinfo{author}{G.~v.~d. Driessche}, \bibinfo{author}{J.-B. Lespiau},
  \bibinfo{author}{B.~Damoc}, \bibinfo{author}{A.~Clark},
  \bibinfo{author}{D.~d.~L. Casas}, \bibinfo{author}{A.~Guy},
  \bibinfo{author}{J.~Menick}, \bibinfo{author}{R.~Ring},
  \bibinfo{author}{T.~Hennigan}, \bibinfo{author}{S.~Huang},
  \bibinfo{author}{L.~Maggiore}, \bibinfo{author}{C.~Jones},
  \bibinfo{author}{A.~Cassirer}, \bibinfo{author}{A.~Brock},
  \bibinfo{author}{M.~Paganini}, \bibinfo{author}{G.~Irving},
  \bibinfo{author}{O.~Vinyals}, \bibinfo{author}{S.~Osindero},
  \bibinfo{author}{K.~Simonyan}, \bibinfo{author}{J.~W. Rae},
  \bibinfo{author}{E.~Elsen}, \bibinfo{author}{L.~Sifre},
  \bibinfo{title}{Improving language models by retrieving from trillions of
  tokens}, \bibinfo{year}{2021}. \URLprefix
  \url{https://arxiv.org/abs/2112.04426}.
  \DOIprefix\doi{10.48550/ARXIV.2112.04426}.
\bibitem[{Baltrušaitis et~al.(2017)Baltrušaitis, Ahuja, and
  Morency}]{survey0}
\bibinfo{author}{T.~Baltrušaitis}, \bibinfo{author}{C.~Ahuja},
  \bibinfo{author}{L.-P. Morency}, \bibinfo{title}{Multimodal machine learning:
  A survey and taxonomy}, \bibinfo{year}{2017}.
  \href{http://arxiv.org/abs/1705.09406}{{\tt arXiv:1705.09406}}.
\bibitem[{Kafle et~al.(2019)Kafle, Shrestha, and Kanan}]{survey1}
\bibinfo{author}{K.~Kafle}, \bibinfo{author}{R.~Shrestha},
  \bibinfo{author}{C.~Kanan}, \bibinfo{title}{Challenges and prospects in
  vision and language research}, \bibinfo{year}{2019}.
  \href{http://arxiv.org/abs/1904.09317}{{\tt arXiv:1904.09317}}.
\bibitem[{Guo et~al.(2019)Guo, Wang, and Wang}]{survey2}
\bibinfo{author}{W.~Guo}, \bibinfo{author}{J.~Wang}, \bibinfo{author}{S.~Wang},
\newblock \bibinfo{title}{Deep multimodal representation learning: A survey},
\newblock \bibinfo{journal}{IEEE Access} \bibinfo{volume}{7}
  (\bibinfo{year}{2019}) \bibinfo{pages}{63373--63394}.
  \DOIprefix\doi{10.1109/ACCESS.2019.2916887}.
\bibitem[{Mogadala et~al.(2021)Mogadala, Kalimuthu, and Klakow}]{survey3}
\bibinfo{author}{A.~Mogadala}, \bibinfo{author}{M.~Kalimuthu},
  \bibinfo{author}{D.~Klakow},
\newblock \bibinfo{title}{Trends in integration of vision and language
  research: A survey of tasks, datasets, and methods},
\newblock \bibinfo{journal}{Journal of Artificial Intelligence Research}
  \bibinfo{volume}{71} (\bibinfo{year}{2021}) \bibinfo{pages}{1183–1317}.
  \URLprefix \url{http://dx.doi.org/10.1613/jair.1.11688}.
  \DOIprefix\doi{10.1613/jair.1.11688}.
\bibitem[{Zhang et~al.(2020)Zhang, Yang, He, and Deng}]{survey4}
\bibinfo{author}{C.~Zhang}, \bibinfo{author}{Z.~Yang}, \bibinfo{author}{X.~He},
  \bibinfo{author}{L.~Deng},
\newblock \bibinfo{title}{Multimodal intelligence: Representation learning,
  information fusion, and applications},
\newblock \bibinfo{journal}{IEEE Journal of Selected Topics in Signal
  Processing} \bibinfo{volume}{14} (\bibinfo{year}{2020})
  \bibinfo{pages}{478–493}. \URLprefix
  \url{http://dx.doi.org/10.1109/JSTSP.2020.2987728}.
  \DOIprefix\doi{10.1109/jstsp.2020.2987728}.
\bibitem[{Uppal et~al.(2020)Uppal, Bhagat, Hazarika, Majumdar, Poria,
  Zimmermann, and Zadeh}]{survey5}
\bibinfo{author}{S.~Uppal}, \bibinfo{author}{S.~Bhagat},
  \bibinfo{author}{D.~Hazarika}, \bibinfo{author}{N.~Majumdar},
  \bibinfo{author}{S.~Poria}, \bibinfo{author}{R.~Zimmermann},
  \bibinfo{author}{A.~Zadeh}, \bibinfo{title}{Multimodal research in vision and
  language: A review of current and emerging trends}, \bibinfo{year}{2020}.
  \href{http://arxiv.org/abs/2010.09522}{{\tt arXiv:2010.09522}}.
\bibitem[{Cao et~al.(2020)Cao, Gan, Cheng, Yu, Chen, and Liu}]{survey6}
\bibinfo{author}{J.~Cao}, \bibinfo{author}{Z.~Gan}, \bibinfo{author}{Y.~Cheng},
  \bibinfo{author}{L.~Yu}, \bibinfo{author}{Y.-C. Chen},
  \bibinfo{author}{J.~Liu},
\newblock \bibinfo{title}{Behind the scene: Revealing the secrets of
  pre-trained vision-and-language models},
\newblock in: \bibinfo{booktitle}{ECCV}, \bibinfo{year}{2020}.
\bibitem[{Lymperaiou and Stamou(2022)}]{kvl-survey}
\bibinfo{author}{M.~Lymperaiou}, \bibinfo{author}{G.~Stamou}, \bibinfo{title}{A
  survey on knowledge-enhanced multimodal learning}, \bibinfo{year}{2022}.
  \URLprefix \url{https://arxiv.org/abs/2211.12328}.
  \DOIprefix\doi{10.48550/ARXIV.2211.12328}.
\bibitem[{Ilievski et~al.(2021)Ilievski, Oltramari, Ma, Zhang, McGuinness, and
  Szekely}]{commonsense-dimensions}
\bibinfo{author}{F.~Ilievski}, \bibinfo{author}{A.~Oltramari},
  \bibinfo{author}{K.~Ma}, \bibinfo{author}{B.~Zhang}, \bibinfo{author}{D.~L.
  McGuinness}, \bibinfo{author}{P.~Szekely},
\newblock \bibinfo{title}{Dimensions of commonsense knowledge}
  (\bibinfo{year}{2021}). \URLprefix \url{https://arxiv.org/abs/2101.04640}.
  \DOIprefix\doi{10.48550/ARXIV.2101.04640}.
\bibitem[{Tanon et~al.(2020)Tanon, Weikum, and Suchanek}]{yago4}
\bibinfo{author}{T.~Tanon}, \bibinfo{author}{G.~Weikum},
  \bibinfo{author}{F.~Suchanek},
\newblock \bibinfo{title}{Yago 4: A reason-able knowledge base}
  (\bibinfo{year}{2020}) \bibinfo{pages}{583--596}.
  \DOIprefix\doi{10.1007/978-3-030-49461-2_34}.
\bibitem[{Zheng and Zhang(2021)}]{kg-query}
\bibinfo{author}{W.~Zheng}, \bibinfo{author}{M.~Zhang},
\newblock \bibinfo{title}{Automated query graph generation for querying
  knowledge graphs},
\newblock in: \bibinfo{booktitle}{Proceedings of the 30th ACM International
  Conference on Information \& Knowledge Management}, CIKM '21,
  \bibinfo{publisher}{Association for Computing Machinery},
  \bibinfo{address}{New York, NY, USA}, \bibinfo{year}{2021}, p.
  \bibinfo{pages}{2698–2707}. \URLprefix
  \url{https://doi.org/10.1145/3459637.3482235}.
  \DOIprefix\doi{10.1145/3459637.3482235}.
\bibitem[{Liu et~al.(2023)Liu, Yuan, Fu, Jiang, Hayashi, and
  Neubig}]{prompt-survey}
\bibinfo{author}{P.~Liu}, \bibinfo{author}{W.~Yuan}, \bibinfo{author}{J.~Fu},
  \bibinfo{author}{Z.~Jiang}, \bibinfo{author}{H.~Hayashi},
  \bibinfo{author}{G.~Neubig},
\newblock \bibinfo{title}{Pre-train, prompt, and predict: A systematic survey
  of prompting methods in natural language processing},
\newblock \bibinfo{journal}{ACM Comput. Surv.} \bibinfo{volume}{55}
  (\bibinfo{year}{2023}). \URLprefix \url{https://doi.org/10.1145/3560815}.
  \DOIprefix\doi{10.1145/3560815}.
\bibitem[{Adolphs et~al.(2021)Adolphs, Dhuliawala, and Hofmann}]{prompt-how}
\bibinfo{author}{L.~Adolphs}, \bibinfo{author}{S.~Dhuliawala},
  \bibinfo{author}{T.~Hofmann}, \bibinfo{title}{How to query language models?},
  \bibinfo{year}{2021}. \URLprefix \url{https://arxiv.org/abs/2108.01928}.
  \DOIprefix\doi{10.48550/ARXIV.2108.01928}.
\bibitem[{Jiang et~al.(2019)Jiang, Xu, Araki, and Neubig}]{prompt-how2}
\bibinfo{author}{Z.~Jiang}, \bibinfo{author}{F.~F. Xu},
  \bibinfo{author}{J.~Araki}, \bibinfo{author}{G.~Neubig}, \bibinfo{title}{How
  can we know what language models know?}, \bibinfo{year}{2019}. \URLprefix
  \url{https://arxiv.org/abs/1911.12543}.
  \DOIprefix\doi{10.48550/ARXIV.1911.12543}.
\bibitem[{Haviv et~al.(2021)Haviv, Berant, and Globerson}]{bertese}
\bibinfo{author}{A.~Haviv}, \bibinfo{author}{J.~Berant},
  \bibinfo{author}{A.~Globerson}, \bibinfo{title}{Bertese: Learning to speak to
  bert}, \bibinfo{year}{2021}. \URLprefix
  \url{https://arxiv.org/abs/2103.05327}.
  \DOIprefix\doi{10.48550/ARXIV.2103.05327}.
\bibitem[{Gao et~al.(2021)Gao, Fisch, and Chen}]{prompt-gen}
\bibinfo{author}{T.~Gao}, \bibinfo{author}{A.~Fisch},
  \bibinfo{author}{D.~Chen},
\newblock \bibinfo{title}{Making pre-trained language models better few-shot
  learners},
\newblock in: \bibinfo{booktitle}{Proceedings of the 59th Annual Meeting of the
  Association for Computational Linguistics and the 11th International Joint
  Conference on Natural Language Processing (Volume 1: Long Papers)},
  \bibinfo{publisher}{Association for Computational Linguistics},
  \bibinfo{address}{Online}, \bibinfo{year}{2021}, pp.
  \bibinfo{pages}{3816--3830}. \URLprefix
  \url{https://aclanthology.org/2021.acl-long.295}.
  \DOIprefix\doi{10.18653/v1/2021.acl-long.295}.
\bibitem[{Guo et~al.(2021)Guo, Tan, Liu, Xing, and Hu}]{rl}
\bibinfo{author}{H.~Guo}, \bibinfo{author}{B.~Tan}, \bibinfo{author}{Z.~Liu},
  \bibinfo{author}{E.~P. Xing}, \bibinfo{author}{Z.~Hu},
  \bibinfo{title}{Efficient (soft) q-learning for text generation with limited
  good data}, \bibinfo{year}{2021}. \URLprefix
  \url{https://arxiv.org/abs/2106.07704}.
  \DOIprefix\doi{10.48550/ARXIV.2106.07704}.
\bibitem[{Li and Liang(2021)}]{prefix-tuning}
\bibinfo{author}{X.~L. Li}, \bibinfo{author}{P.~Liang},
\newblock \bibinfo{title}{Prefix-tuning: Optimizing continuous prompts for
  generation},
\newblock in: \bibinfo{booktitle}{Proceedings of the 59th Annual Meeting of the
  Association for Computational Linguistics and the 11th International Joint
  Conference on Natural Language Processing (Volume 1: Long Papers)},
  \bibinfo{publisher}{Association for Computational Linguistics},
  \bibinfo{address}{Online}, \bibinfo{year}{2021}, pp.
  \bibinfo{pages}{4582--4597}. \URLprefix
  \url{https://aclanthology.org/2021.acl-long.353}.
  \DOIprefix\doi{10.18653/v1/2021.acl-long.353}.
\bibitem[{Zhong et~al.(2021)Zhong, Friedman, and Chen}]{factual-probing}
\bibinfo{author}{Z.~Zhong}, \bibinfo{author}{D.~Friedman},
  \bibinfo{author}{D.~Chen}, \bibinfo{title}{Factual probing is [mask]:
  Learning vs. learning to recall}, \bibinfo{year}{2021}. \URLprefix
  \url{https://arxiv.org/abs/2104.05240}.
  \DOIprefix\doi{10.48550/ARXIV.2104.05240}.
\bibitem[{Huang and Chang(2022)}]{reasoning}
\bibinfo{author}{J.~Huang}, \bibinfo{author}{K.~C.-C. Chang},
\newblock \bibinfo{title}{Towards reasoning in large language models: A
  survey},
\newblock \bibinfo{journal}{ArXiv} \bibinfo{volume}{abs/2212.10403}
  (\bibinfo{year}{2022}).
\bibitem[{Wei et~al.(2022)Wei, Tay, Bommasani, Raffel, Zoph, Borgeaud,
  Yogatama, Bosma, Zhou, Metzler, Chi, Hashimoto, Vinyals, Liang, Dean, and
  Fedus}]{emerging}
\bibinfo{author}{J.~Wei}, \bibinfo{author}{Y.~Tay},
  \bibinfo{author}{R.~Bommasani}, \bibinfo{author}{C.~Raffel},
  \bibinfo{author}{B.~Zoph}, \bibinfo{author}{S.~Borgeaud},
  \bibinfo{author}{D.~Yogatama}, \bibinfo{author}{M.~Bosma},
  \bibinfo{author}{D.~Zhou}, \bibinfo{author}{D.~Metzler},
  \bibinfo{author}{E.~H. Chi}, \bibinfo{author}{T.~Hashimoto},
  \bibinfo{author}{O.~Vinyals}, \bibinfo{author}{P.~Liang},
  \bibinfo{author}{J.~Dean}, \bibinfo{author}{W.~Fedus},
  \bibinfo{title}{Emergent abilities of large language models},
  \bibinfo{year}{2022}. \URLprefix \url{https://arxiv.org/abs/2206.07682}.
  \DOIprefix\doi{10.48550/ARXIV.2206.07682}.
\bibitem[{Kojima et~al.(2022)Kojima, Gu, Reid, Matsuo, and Iwasawa}]{cot}
\bibinfo{author}{T.~Kojima}, \bibinfo{author}{S.~S. Gu},
  \bibinfo{author}{M.~Reid}, \bibinfo{author}{Y.~Matsuo},
  \bibinfo{author}{Y.~Iwasawa}, \bibinfo{title}{Large language models are
  zero-shot reasoners}, \bibinfo{year}{2022}. \URLprefix
  \url{https://arxiv.org/abs/2205.11916}.
  \DOIprefix\doi{10.48550/ARXIV.2205.11916}.
\bibitem[{Wei et~al.(2022)Wei, Wang, Schuurmans, Bosma, hsin Chi, Le, and
  Zhou}]{cot-reasoning}
\bibinfo{author}{J.~Wei}, \bibinfo{author}{X.~Wang},
  \bibinfo{author}{D.~Schuurmans}, \bibinfo{author}{M.~Bosma},
  \bibinfo{author}{E.~H. hsin Chi}, \bibinfo{author}{Q.~Le},
  \bibinfo{author}{D.~Zhou},
\newblock \bibinfo{title}{Chain of thought prompting elicits reasoning in large
  language models},
\newblock \bibinfo{journal}{ArXiv} \bibinfo{volume}{abs/2201.11903}
  (\bibinfo{year}{2022}).
\bibitem[{Amini et~al.(2019)Amini, Gabriel, Lin, Koncel-Kedziorski, Choi, and
  Hajishirzi}]{mathqa}
\bibinfo{author}{A.~Amini}, \bibinfo{author}{S.~Gabriel},
  \bibinfo{author}{S.~Lin}, \bibinfo{author}{R.~Koncel-Kedziorski},
  \bibinfo{author}{Y.~Choi}, \bibinfo{author}{H.~Hajishirzi},
\newblock \bibinfo{title}{{M}ath{QA}: Towards interpretable math word problem
  solving with operation-based formalisms},
\newblock in: \bibinfo{booktitle}{Proceedings of the 2019 Conference of the
  North {A}merican Chapter of the Association for Computational Linguistics:
  Human Language Technologies, Volume 1 (Long and Short Papers)},
  \bibinfo{publisher}{Association for Computational Linguistics},
  \bibinfo{address}{Minneapolis, Minnesota}, \bibinfo{year}{2019}, pp.
  \bibinfo{pages}{2357--2367}. \URLprefix
  \url{https://aclanthology.org/N19-1245}.
  \DOIprefix\doi{10.18653/v1/N19-1245}.
\bibitem[{Bhargava and Ng(2022)}]{commonsense-survey}
\bibinfo{author}{P.~Bhargava}, \bibinfo{author}{V.~Ng},
  \bibinfo{title}{Commonsense knowledge reasoning and generation with
  pre-trained language models: A survey}, \bibinfo{year}{2022}. \URLprefix
  \url{https://arxiv.org/abs/2201.12438}.
  \DOIprefix\doi{10.48550/ARXIV.2201.12438}.
\bibitem[{Wu et~al.(2016)Wu, Wang, Shen, Dick, and van~den Hengel}]{vqa1}
\bibinfo{author}{Q.~Wu}, \bibinfo{author}{P.~Wang}, \bibinfo{author}{C.~Shen},
  \bibinfo{author}{A.~R. Dick}, \bibinfo{author}{A.~van~den Hengel},
\newblock \bibinfo{title}{Ask me anything: Free-form visual question answering
  based on knowledge from external sources},
\newblock \bibinfo{journal}{2016 IEEE Conference on Computer Vision and Pattern
  Recognition (CVPR)}  (\bibinfo{year}{2016}) \bibinfo{pages}{4622--4630}.
\bibitem[{Wang et~al.(2018)Wang, Wu, Shen, Dick, and van~den Hengel}]{fvqa}
\bibinfo{author}{P.~Wang}, \bibinfo{author}{Q.~Wu}, \bibinfo{author}{C.~Shen},
  \bibinfo{author}{A.~R. Dick}, \bibinfo{author}{A.~van~den Hengel},
\newblock \bibinfo{title}{Fvqa: Fact-based visual question answering},
\newblock \bibinfo{journal}{IEEE Transactions on Pattern Analysis and Machine
  Intelligence} \bibinfo{volume}{40} (\bibinfo{year}{2018})
  \bibinfo{pages}{2413--2427}.
\bibitem[{Singh et~al.(2019)Singh, Mishra, Shekhar, and Chakraborty}]{vqa6}
\bibinfo{author}{A.~K. Singh}, \bibinfo{author}{A.~Mishra},
  \bibinfo{author}{S.~Shekhar}, \bibinfo{author}{A.~Chakraborty},
\newblock \bibinfo{title}{From strings to things: Knowledge-enabled vqa model
  that can read and reason},
\newblock \bibinfo{journal}{2019 IEEE/CVF International Conference on Computer
  Vision (ICCV)}  (\bibinfo{year}{2019}) \bibinfo{pages}{4601--4611}.
\bibitem[{Yu et~al.(2020)Yu, Zhu, Wang, Zhang, Hu, and Tan}]{vqa8}
\bibinfo{author}{J.~Yu}, \bibinfo{author}{Z.~Zhu}, \bibinfo{author}{Y.~Wang},
  \bibinfo{author}{W.~Zhang}, \bibinfo{author}{Y.~Hu},
  \bibinfo{author}{J.~Tan},
\newblock \bibinfo{title}{Cross-modal knowledge reasoning for knowledge-based
  visual question answering},
\newblock \bibinfo{journal}{ArXiv} \bibinfo{volume}{abs/2009.00145}
  (\bibinfo{year}{2020}).
\bibitem[{Chen et~al.(2021)Chen, Chen, Geng, Pan, Yuan, and Chen}]{vqa21}
\bibinfo{author}{Z.~Chen}, \bibinfo{author}{J.~Chen},
  \bibinfo{author}{Y.~Geng}, \bibinfo{author}{J.~Z. Pan},
  \bibinfo{author}{Z.~Yuan}, \bibinfo{author}{H.~Chen},
\newblock \bibinfo{title}{Zero-shot visual question answering using knowledge
  graph},
\newblock in: \bibinfo{booktitle}{SEMWEB}, \bibinfo{year}{2021}.
\bibitem[{Zhu et~al.(2015)Zhu, Zhang, Ré, and Fei-Fei}]{vqa0}
\bibinfo{author}{Y.~Zhu}, \bibinfo{author}{C.~Zhang}, \bibinfo{author}{C.~Ré},
  \bibinfo{author}{L.~Fei-Fei}, \bibinfo{title}{Building a large-scale
  multimodal knowledge base system for answering visual queries},
  \bibinfo{year}{2015}. \href{http://arxiv.org/abs/1507.05670}{{\tt
  arXiv:1507.05670}}.
\bibitem[{Wang et~al.(2017)Wang, Wu, Shen, Dick, and van~den Hengel}]{vqa2}
\bibinfo{author}{P.~Wang}, \bibinfo{author}{Q.~Wu}, \bibinfo{author}{C.~Shen},
  \bibinfo{author}{A.~R. Dick}, \bibinfo{author}{A.~van~den Hengel},
\newblock \bibinfo{title}{Explicit knowledge-based reasoning for visual
  question answering},
\newblock in: \bibinfo{booktitle}{IJCAI}, \bibinfo{year}{2017}.
\bibitem[{Narasimhan and Schwing(2018)}]{vqa3}
\bibinfo{author}{M.~Narasimhan}, \bibinfo{author}{A.~G. Schwing},
\newblock \bibinfo{title}{Straight to the facts: Learning knowledge base
  retrieval for factual visual question answering},
\newblock \bibinfo{journal}{ArXiv} \bibinfo{volume}{abs/1809.01124}
  (\bibinfo{year}{2018}).
\bibitem[{Pennington et~al.(2014)Pennington, Socher, and Manning}]{glove}
\bibinfo{author}{J.~Pennington}, \bibinfo{author}{R.~Socher},
  \bibinfo{author}{C.~Manning},
\newblock \bibinfo{title}{{G}lo{V}e: Global vectors for word representation},
\newblock in: \bibinfo{booktitle}{Proceedings of the 2014 Conference on
  Empirical Methods in Natural Language Processing ({EMNLP})},
  \bibinfo{publisher}{Association for Computational Linguistics},
  \bibinfo{address}{Doha, Qatar}, \bibinfo{year}{2014}, pp.
  \bibinfo{pages}{1532--1543}. \URLprefix
  \url{https://aclanthology.org/D14-1162}. \DOIprefix\doi{10.3115/v1/D14-1162}.
\bibitem[{Mikolov et~al.(2013)Mikolov, Sutskever, Chen, Corrado, and
  Dean}]{word2vec}
\bibinfo{author}{T.~Mikolov}, \bibinfo{author}{I.~Sutskever},
  \bibinfo{author}{K.~Chen}, \bibinfo{author}{G.~Corrado},
  \bibinfo{author}{J.~Dean}, \bibinfo{title}{Distributed representations of
  words and phrases and their compositionality}, \bibinfo{year}{2013}.
  \href{http://arxiv.org/abs/1310.4546}{{\tt arXiv:1310.4546}}.
\bibitem[{Ziaeefard and L{\'e}cu{\'e}(2020)}]{vqa9}
\bibinfo{author}{M.~Ziaeefard}, \bibinfo{author}{F.~L{\'e}cu{\'e}},
\newblock \bibinfo{title}{Towards knowledge-augmented visual question
  answering},
\newblock in: \bibinfo{booktitle}{COLING}, \bibinfo{year}{2020}.
\bibitem[{Salaberria et~al.(2021)Salaberria, Azkune, de~Lacalle, Etxabe, and
  Agirre}]{vqa19}
\bibinfo{author}{A.~Salaberria}, \bibinfo{author}{G.~Azkune},
  \bibinfo{author}{O.~L. de~Lacalle}, \bibinfo{author}{A.~S. Etxabe},
  \bibinfo{author}{E.~Agirre},
\newblock \bibinfo{title}{Image captioning for effective use of language models
  in knowledge-based visual question answering},
\newblock \bibinfo{journal}{ArXiv} \bibinfo{volume}{abs/2109.08029}
  (\bibinfo{year}{2021}).
\bibitem[{Gard{\`e}res et~al.(2020)Gard{\`e}res, Ziaeefard, Abeloos, and
  L{\'e}cu{\'e}}]{vqa7}
\bibinfo{author}{F.~Gard{\`e}res}, \bibinfo{author}{M.~Ziaeefard},
  \bibinfo{author}{B.~Abeloos}, \bibinfo{author}{F.~L{\'e}cu{\'e}},
\newblock \bibinfo{title}{Conceptbert: Concept-aware representation for visual
  question answering},
\newblock in: \bibinfo{booktitle}{FINDINGS}, \bibinfo{year}{2020}.
\bibitem[{Wu et~al.(2021)Wu, Lu, Sabharwal, and Mottaghi}]{vqa14}
\bibinfo{author}{J.~Wu}, \bibinfo{author}{J.~Lu},
  \bibinfo{author}{A.~Sabharwal}, \bibinfo{author}{R.~Mottaghi},
\newblock \bibinfo{title}{Multi-modal answer validation for knowledge-based
  vqa},
\newblock \bibinfo{journal}{ArXiv} \bibinfo{volume}{abs/2103.12248}
  (\bibinfo{year}{2021}).
\bibitem[{Qu et~al.(2021)Qu, Zamani, Yang, Croft, and Learned-Miller}]{vqa17}
\bibinfo{author}{C.~Qu}, \bibinfo{author}{H.~Zamani},
  \bibinfo{author}{L.~Yang}, \bibinfo{author}{W.~B. Croft},
  \bibinfo{author}{E.~G. Learned-Miller},
\newblock \bibinfo{title}{Passage retrieval for outside-knowledge visual
  question answering},
\newblock \bibinfo{journal}{Proceedings of the 44th International ACM SIGIR
  Conference on Research and Development in Information Retrieval}
  (\bibinfo{year}{2021}).
\bibitem[{Gao et~al.(2022)Gao, Ping, Thattai, Reganti, Wu, and
  Natarajan}]{vqa22}
\bibinfo{author}{F.~Gao}, \bibinfo{author}{Q.~Ping},
  \bibinfo{author}{G.~Thattai}, \bibinfo{author}{A.~N. Reganti},
  \bibinfo{author}{Y.~Wu}, \bibinfo{author}{P.~Natarajan},
\newblock \bibinfo{title}{Transform-retrieve-generate: Natural language-centric
  outside-knowledge visual question answering},
\newblock \bibinfo{journal}{2022 IEEE/CVF Conference on Computer Vision and
  Pattern Recognition (CVPR)}  (\bibinfo{year}{2022})
  \bibinfo{pages}{5057--5067}.
\bibitem[{Raffel et~al.(2020)Raffel, Shazeer, Roberts, Lee, Narang, Matena,
  Zhou, Li, and Liu}]{t5}
\bibinfo{author}{C.~Raffel}, \bibinfo{author}{N.~Shazeer},
  \bibinfo{author}{A.~Roberts}, \bibinfo{author}{K.~Lee},
  \bibinfo{author}{S.~Narang}, \bibinfo{author}{M.~Matena},
  \bibinfo{author}{Y.~Zhou}, \bibinfo{author}{W.~Li}, \bibinfo{author}{P.~J.
  Liu},
\newblock \bibinfo{title}{Exploring the limits of transfer learning with a
  unified text-to-text transformer},
\newblock \bibinfo{journal}{Journal of Machine Learning Research}
  \bibinfo{volume}{21} (\bibinfo{year}{2020}) \bibinfo{pages}{1--67}.
  \URLprefix \url{http://jmlr.org/papers/v21/20-074.html}.
\bibitem[{Lin and Byrne(2022)}]{vqa23}
\bibinfo{author}{W.~Lin}, \bibinfo{author}{B.~Byrne},
\newblock \bibinfo{title}{Retrieval augmented visual question answering with
  outside knowledge},
\newblock in: \bibinfo{booktitle}{Conference on Empirical Methods in Natural
  Language Processing}, \bibinfo{year}{2022}.
\bibitem[{Wu and Mooney(2022)}]{vqa24}
\bibinfo{author}{J.~Wu}, \bibinfo{author}{R.~J. Mooney},
\newblock \bibinfo{title}{Entity-focused dense passage retrieval for
  outside-knowledge visual question answering},
\newblock in: \bibinfo{booktitle}{Conference on Empirical Methods in Natural
  Language Processing}, \bibinfo{year}{2022}.
\bibitem[{Tandon et~al.(2014)Tandon, de~Melo, and Weikum}]{webchild}
\bibinfo{author}{N.~Tandon}, \bibinfo{author}{G.~de~Melo},
  \bibinfo{author}{G.~Weikum},
\newblock \bibinfo{title}{Acquiring comparative commonsense knowledge from the
  web},
\newblock \bibinfo{journal}{Proceedings of the National Conference on
  Artificial Intelligence} \bibinfo{volume}{1} (\bibinfo{year}{2014})
  \bibinfo{pages}{166--172}.
\bibitem[{Bhakthavatsalam et~al.(2020)Bhakthavatsalam, Richardson, Tandon, and
  Clark}]{haspart}
\bibinfo{author}{S.~Bhakthavatsalam}, \bibinfo{author}{K.~Richardson},
  \bibinfo{author}{N.~Tandon}, \bibinfo{author}{P.~Clark}, \bibinfo{title}{Do
  dogs have whiskers? a new knowledge base of haspart relations},
  \bibinfo{year}{2020}. \href{http://arxiv.org/abs/2006.07510}{{\tt
  arXiv:2006.07510}}.
\bibitem[{Chen et~al.(2022)Chen, Huang, Chen, Geng, Fang, Pan, Zhang, and
  Zhang}]{vqa25}
\bibinfo{author}{Z.~Chen}, \bibinfo{author}{Y.~Huang},
  \bibinfo{author}{J.~Chen}, \bibinfo{author}{Y.~Geng},
  \bibinfo{author}{Y.~Fang}, \bibinfo{author}{J.~Z. Pan},
  \bibinfo{author}{N.~Zhang}, \bibinfo{author}{W.~Zhang},
\newblock \bibinfo{title}{Lako: Knowledge-driven visual question answering via
  late knowledge-to-text injection},
\newblock \bibinfo{journal}{Proceedings of the 11th International Joint
  Conference on Knowledge Graphs}  (\bibinfo{year}{2022}).
\bibitem[{Lerner et~al.(2023)Lerner, Ferret, and Guinaudeau}]{vqa26}
\bibinfo{author}{P.~Lerner}, \bibinfo{author}{O.~Ferret},
  \bibinfo{author}{C.~Guinaudeau},
\newblock \bibinfo{title}{Multimodal inverse cloze task for knowledge-based
  visual question answering},
\newblock \bibinfo{journal}{ArXiv} \bibinfo{volume}{abs/2301.04366}
  (\bibinfo{year}{2023}).
\bibitem[{Yang et~al.(2021)Yang, Gan, Wang, Hu, Lu, Liu, and Wang}]{vqa15}
\bibinfo{author}{Z.~Yang}, \bibinfo{author}{Z.~Gan}, \bibinfo{author}{J.~Wang},
  \bibinfo{author}{X.~Hu}, \bibinfo{author}{Y.~Lu}, \bibinfo{author}{Z.~Liu},
  \bibinfo{author}{L.~Wang},
\newblock \bibinfo{title}{An empirical study of gpt-3 for few-shot
  knowledge-based vqa},
\newblock \bibinfo{journal}{ArXiv} \bibinfo{volume}{abs/2109.05014}
  (\bibinfo{year}{2021}).
\bibitem[{Tiong et~al.(2022)Tiong, Li, Li, Savarese, and Hoi}]{vqa-lm1}
\bibinfo{author}{A.~M.~H. Tiong}, \bibinfo{author}{J.~Li},
  \bibinfo{author}{B.~Li}, \bibinfo{author}{S.~Savarese},
  \bibinfo{author}{S.~C.~H. Hoi},
\newblock \bibinfo{title}{Plug-and-play vqa: Zero-shot vqa by conjoining large
  pretrained models with zero training},
\newblock in: \bibinfo{booktitle}{Conference on Empirical Methods in Natural
  Language Processing}, \bibinfo{year}{2022}.
\bibitem[{Guo et~al.(2022)Guo, Li, Li, Tiong, Li, Tao, and Hoi}]{vqa-lm2}
\bibinfo{author}{J.~Guo}, \bibinfo{author}{J.~Li}, \bibinfo{author}{D.~Li},
  \bibinfo{author}{A.~M.~H. Tiong}, \bibinfo{author}{B.~Li},
  \bibinfo{author}{D.~Tao}, \bibinfo{author}{S.~Hoi},
\newblock \bibinfo{title}{From images to textual prompts: Zero-shot vqa with
  frozen large language models},
\newblock \bibinfo{journal}{ArXiv} \bibinfo{volume}{abs/2212.10846}
  (\bibinfo{year}{2022}).
\bibitem[{Chen et~al.(2023)Chen, Zhou, Shen, Hong, Zhang, and Gan}]{vqa-lm5}
\bibinfo{author}{Z.~Chen}, \bibinfo{author}{Q.~Zhou},
  \bibinfo{author}{Y.~Shen}, \bibinfo{author}{Y.~Hong},
  \bibinfo{author}{H.~Zhang}, \bibinfo{author}{C.~Gan},
\newblock \bibinfo{title}{See, think, confirm: Interactive prompting between
  vision and language models for knowledge-based visual reasoning},
\newblock \bibinfo{journal}{ArXiv} \bibinfo{volume}{abs/2301.05226}
  (\bibinfo{year}{2023}).
\bibitem[{Jin et~al.(2021)Jin, Cheng, Shen, Chen, and Ren}]{vqa-lm4}
\bibinfo{author}{W.~Jin}, \bibinfo{author}{Y.~Cheng},
  \bibinfo{author}{Y.~Shen}, \bibinfo{author}{W.~Chen},
  \bibinfo{author}{X.~Ren}, \bibinfo{title}{A good prompt is worth millions of
  parameters: Low-resource prompt-based learning for vision-language models},
  \bibinfo{year}{2021}. \URLprefix \url{https://arxiv.org/abs/2110.08484}.
  \DOIprefix\doi{10.48550/ARXIV.2110.08484}.
\bibitem[{Marino et~al.(2021)Marino, Chen, Parikh, Gupta, and Rohrbach}]{vqa16}
\bibinfo{author}{K.~Marino}, \bibinfo{author}{X.~Chen},
  \bibinfo{author}{D.~Parikh}, \bibinfo{author}{A.~K. Gupta},
  \bibinfo{author}{M.~Rohrbach},
\newblock \bibinfo{title}{Krisp: Integrating implicit and symbolic knowledge
  for open-domain knowledge-based vqa},
\newblock \bibinfo{journal}{2021 IEEE/CVF Conference on Computer Vision and
  Pattern Recognition (CVPR)}  (\bibinfo{year}{2021})
  \bibinfo{pages}{14106--14116}.
\bibitem[{Krishna et~al.(2016)Krishna, Zhu, Groth, Johnson, Hata, Kravitz,
  Chen, Kalantidis, Li, Shamma, Bernstein, and Li}]{visualgenome}
\bibinfo{author}{R.~Krishna}, \bibinfo{author}{Y.~Zhu},
  \bibinfo{author}{O.~Groth}, \bibinfo{author}{J.~Johnson},
  \bibinfo{author}{K.~Hata}, \bibinfo{author}{J.~Kravitz},
  \bibinfo{author}{S.~Chen}, \bibinfo{author}{Y.~Kalantidis},
  \bibinfo{author}{L.-J. Li}, \bibinfo{author}{D.~A. Shamma},
  \bibinfo{author}{M.~S. Bernstein}, \bibinfo{author}{F.-F. Li},
  \bibinfo{title}{Visual genome: Connecting language and vision using
  crowdsourced dense image annotations}, \bibinfo{year}{2016}.
  \href{http://arxiv.org/abs/1602.07332}{{\tt arXiv:1602.07332}}.
\bibitem[{Lin et~al.(2022)Lin, Xie, Chen, Xu, Zhu, and Yuan}]{revive}
\bibinfo{author}{Y.~Lin}, \bibinfo{author}{Y.~Xie}, \bibinfo{author}{D.~Chen},
  \bibinfo{author}{Y.~Xu}, \bibinfo{author}{C.~Zhu}, \bibinfo{author}{L.~Yuan},
  \bibinfo{title}{Revive: Regional visual representation matters in
  knowledge-based visual question answering}, \bibinfo{year}{2022}. \URLprefix
  \url{https://arxiv.org/abs/2206.01201}.
  \DOIprefix\doi{10.48550/ARXIV.2206.01201}.
\bibitem[{Gui et~al.(2022)Gui, Wang, Huang, Hauptmann, Bisk, and Gao}]{vqa-lm0}
\bibinfo{author}{L.~Gui}, \bibinfo{author}{B.~Wang},
  \bibinfo{author}{Q.~Huang}, \bibinfo{author}{A.~Hauptmann},
  \bibinfo{author}{Y.~Bisk}, \bibinfo{author}{J.~Gao},
\newblock \bibinfo{title}{{KAT}: A knowledge augmented transformer for
  vision-and-language},
\newblock in: \bibinfo{booktitle}{Proceedings of the 2022 Conference of the
  North American Chapter of the Association for Computational Linguistics:
  Human Language Technologies}, \bibinfo{publisher}{Association for
  Computational Linguistics}, \bibinfo{address}{Seattle, United States},
  \bibinfo{year}{2022}, pp. \bibinfo{pages}{956--968}. \URLprefix
  \url{https://aclanthology.org/2022.naacl-main.70}.
  \DOIprefix\doi{10.18653/v1/2022.naacl-main.70}.
\bibitem[{Hwang et~al.(2021)Hwang, Bhagavatula, {Le Bras}, Da, Sakaguchi,
  Bosselut, and Choi}]{atomic}
\bibinfo{author}{J.~D. Hwang}, \bibinfo{author}{C.~Bhagavatula},
  \bibinfo{author}{R.~{Le Bras}}, \bibinfo{author}{J.~Da},
  \bibinfo{author}{K.~Sakaguchi}, \bibinfo{author}{A.~Bosselut},
  \bibinfo{author}{Y.~Choi},
\newblock \bibinfo{title}{Comet-atomic 2020: On symbolic and neural commonsense
  knowledge graphs},
\newblock in: \bibinfo{booktitle}{AAAI}, \bibinfo{year}{2021}.
\bibitem[{Zellers et~al.(2018)Zellers, Bisk, Schwartz, and Choi}]{swag}
\bibinfo{author}{R.~Zellers}, \bibinfo{author}{Y.~Bisk},
  \bibinfo{author}{R.~Schwartz}, \bibinfo{author}{Y.~Choi},
  \bibinfo{title}{Swag: A large-scale adversarial dataset for grounded
  commonsense inference}, \bibinfo{year}{2018}. \URLprefix
  \url{https://arxiv.org/abs/1808.05326}.
  \DOIprefix\doi{10.48550/ARXIV.1808.05326}.
\bibitem[{Song et~al.(2021)Song, Ma, Sun, Yang, and Liao}]{vcr6}
\bibinfo{author}{D.~Song}, \bibinfo{author}{S.~Ma}, \bibinfo{author}{Z.~Sun},
  \bibinfo{author}{S.~Yang}, \bibinfo{author}{L.~Liao},
\newblock \bibinfo{title}{Kvl-bert: Knowledge enhanced visual-and-linguistic
  bert for visual commonsense reasoning},
\newblock \bibinfo{journal}{Knowl. Based Syst.} \bibinfo{volume}{230}
  (\bibinfo{year}{2021}) \bibinfo{pages}{107408}.
\bibitem[{Su et~al.(2020)Su, Zhu, Cao, Li, Lu, Wei, and Dai}]{su2020vlbert}
\bibinfo{author}{W.~Su}, \bibinfo{author}{X.~Zhu}, \bibinfo{author}{Y.~Cao},
  \bibinfo{author}{B.~Li}, \bibinfo{author}{L.~Lu}, \bibinfo{author}{F.~Wei},
  \bibinfo{author}{J.~Dai}, \bibinfo{title}{Vl-bert: Pre-training of generic
  visual-linguistic representations}, \bibinfo{year}{2020}.
  \href{http://arxiv.org/abs/1908.08530}{{\tt arXiv:1908.08530}}.
\bibitem[{Lee and Kim(2020)}]{vcr2}
\bibinfo{author}{J.~Lee}, \bibinfo{author}{I.~Kim},
\newblock \bibinfo{title}{Vision–language–knowledge co-embedding for visual
  commonsense reasoning},
\newblock \bibinfo{year}{2020}.
\bibitem[{Ye et~al.(2022)Ye, Xie, Chen, Xu, Yuan, Zhu, and Liao}]{vcr9}
\bibinfo{author}{S.~Ye}, \bibinfo{author}{Y.~Xie}, \bibinfo{author}{D.~Chen},
  \bibinfo{author}{Y.~Xu}, \bibinfo{author}{L.~Yuan}, \bibinfo{author}{C.~Zhu},
  \bibinfo{author}{J.~Liao}, \bibinfo{title}{Improving commonsense in
  vision-language models via knowledge graph riddles}, \bibinfo{year}{2022}.
  \URLprefix \url{https://arxiv.org/abs/2211.16504}.
  \DOIprefix\doi{10.48550/ARXIV.2211.16504}.
\bibitem[{Radford et~al.(2019)Radford, Wu, Child, Luan, Amodei, and
  Sutskever}]{gpt2}
\bibinfo{author}{A.~Radford}, \bibinfo{author}{J.~Wu},
  \bibinfo{author}{R.~Child}, \bibinfo{author}{D.~Luan},
  \bibinfo{author}{D.~Amodei}, \bibinfo{author}{I.~Sutskever},
\newblock \bibinfo{title}{Language models are unsupervised multitask learners},
\newblock \bibinfo{year}{2019}.
\bibitem[{Park et~al.(2020)Park, Bhagavatula, Mottaghi, Farhadi, and
  Choi}]{visualcomet}
\bibinfo{author}{J.~S. Park}, \bibinfo{author}{C.~Bhagavatula},
  \bibinfo{author}{R.~Mottaghi}, \bibinfo{author}{A.~Farhadi},
  \bibinfo{author}{Y.~Choi},
\newblock \bibinfo{title}{Visualcomet: Reasoning about the dynamic context of a
  still image},
\newblock in: \bibinfo{booktitle}{In Proceedings of the European Conference on
  Computer Vision (ECCV)}, \bibinfo{year}{2020}.
\bibitem[{Lu et~al.(2022)Lu, Mishra, Xia, Qiu, Chang, Zhu, Tafjord, Clark, and
  Kalyan}]{vcr10}
\bibinfo{author}{P.~Lu}, \bibinfo{author}{S.~Mishra}, \bibinfo{author}{T.~Xia},
  \bibinfo{author}{L.~Qiu}, \bibinfo{author}{K.-W. Chang},
  \bibinfo{author}{S.-C. Zhu}, \bibinfo{author}{O.~Tafjord},
  \bibinfo{author}{P.~Clark}, \bibinfo{author}{A.~Kalyan},
  \bibinfo{title}{Learn to explain: Multimodal reasoning via thought chains for
  science question answering}, \bibinfo{year}{2022}. \URLprefix
  \url{https://arxiv.org/abs/2209.09513}.
  \DOIprefix\doi{10.48550/ARXIV.2209.09513}.
\bibitem[{Zhang et~al.(2023)Zhang, Zhang, Li, Zhao, Karypis, and Smola}]{vcr11}
\bibinfo{author}{Z.~Zhang}, \bibinfo{author}{A.~Zhang},
  \bibinfo{author}{M.~Li}, \bibinfo{author}{H.~Zhao},
  \bibinfo{author}{G.~Karypis}, \bibinfo{author}{A.~Smola},
  \bibinfo{title}{Multimodal chain-of-thought reasoning in language models},
  \bibinfo{year}{2023}. \URLprefix \url{https://arxiv.org/abs/2302.00923}.
  \DOIprefix\doi{10.48550/ARXIV.2302.00923}.
\bibitem[{Daull et~al.(2023)Daull, Bellot, Bruno, Martin, and
  Murisasco}]{ComplexQA}
\bibinfo{author}{X.~Daull}, \bibinfo{author}{P.~Bellot},
  \bibinfo{author}{E.~Bruno}, \bibinfo{author}{V.~Martin},
  \bibinfo{author}{E.~Murisasco},
\newblock \bibinfo{title}{Complex qa and language models hybrid architectures,
  survey},
\newblock \bibinfo{year}{2023}.
\bibitem[{Zhou et~al.(2019)Zhou, Sun, and Honavar}]{caption1}
\bibinfo{author}{Y.~Zhou}, \bibinfo{author}{Y.~Sun}, \bibinfo{author}{V.~G.
  Honavar},
\newblock \bibinfo{title}{Improving image captioning by leveraging knowledge
  graphs},
\newblock \bibinfo{journal}{2019 IEEE Winter Conference on Applications of
  Computer Vision (WACV)}  (\bibinfo{year}{2019}) \bibinfo{pages}{283--293}.
\bibitem[{Hou et~al.(2019)Hou, Wu, Qi, Zhao, Luo, and Jia}]{caption2}
\bibinfo{author}{J.~Hou}, \bibinfo{author}{X.~Wu}, \bibinfo{author}{Y.~Qi},
  \bibinfo{author}{W.~Zhao}, \bibinfo{author}{J.~Luo},
  \bibinfo{author}{Y.~Jia},
\newblock \bibinfo{title}{Relational reasoning using prior knowledge for visual
  captioning},
\newblock \bibinfo{journal}{ArXiv} \bibinfo{volume}{abs/1906.01290}
  (\bibinfo{year}{2019}).
\bibitem[{Huang et~al.(2020)Huang, Li, Chen, Zhang, and Ma}]{caption3}
\bibinfo{author}{F.~Huang}, \bibinfo{author}{Z.~Li}, \bibinfo{author}{S.~Chen},
  \bibinfo{author}{C.~Zhang}, \bibinfo{author}{H.~Ma},
\newblock \bibinfo{title}{Image captioning with internal and external
  knowledge},
\newblock \bibinfo{journal}{Proceedings of the 29th ACM International
  Conference on Information \& Knowledge Management}  (\bibinfo{year}{2020}).
\bibitem[{Aditya~Mogadala(2020)}]{caption5}
\bibinfo{author}{D.~K. Aditya~Mogadala, Xiaoyu~Shen},
\newblock \bibinfo{title}{Integrating rule-based entity masking into image
  captioning},
\newblock \bibinfo{year}{2020}.
\bibitem[{Zhao et~al.(2021)Zhao, Hu, Wang, Wu, and Luo}]{caption10}
\bibinfo{author}{W.~Zhao}, \bibinfo{author}{Y.~Hu}, \bibinfo{author}{H.~Wang},
  \bibinfo{author}{X.~Wu}, \bibinfo{author}{J.~Luo}, \bibinfo{title}{Boosting
  entity-aware image captioning with multi-modal knowledge graph},
  \bibinfo{year}{2021}. \URLprefix \url{https://arxiv.org/abs/2107.11970}.
  \DOIprefix\doi{10.48550/ARXIV.2107.11970}.
\bibitem[{Xing et~al.(2021)Xing, Shi, Meng, Ma, and Wattenhofer}]{caption8}
\bibinfo{author}{Y.~Xing}, \bibinfo{author}{Z.~Shi}, \bibinfo{author}{Z.~Meng},
  \bibinfo{author}{Y.~Ma}, \bibinfo{author}{R.~Wattenhofer},
\newblock \bibinfo{title}{Km-bart: Knowledge enhanced multimodal bart for
  visual commonsense generation},
\newblock in: \bibinfo{booktitle}{ACL/IJCNLP}, \bibinfo{year}{2021}.
\bibitem[{Lewis et~al.(2019)Lewis, Liu, Goyal, Ghazvininejad, Mohamed, Levy,
  Stoyanov, and Zettlemoyer}]{bart}
\bibinfo{author}{M.~Lewis}, \bibinfo{author}{Y.~Liu},
  \bibinfo{author}{N.~Goyal}, \bibinfo{author}{M.~Ghazvininejad},
  \bibinfo{author}{A.~Mohamed}, \bibinfo{author}{O.~Levy},
  \bibinfo{author}{V.~Stoyanov}, \bibinfo{author}{L.~Zettlemoyer},
  \bibinfo{title}{Bart: Denoising sequence-to-sequence pre-training for natural
  language generation, translation, and comprehension}, \bibinfo{year}{2019}.
  \URLprefix \url{https://arxiv.org/abs/1910.13461}.
  \DOIprefix\doi{10.48550/ARXIV.1910.13461}.
\bibitem[{Nikiforova et~al.(2022)Nikiforova, Deoskar, Paperno, and
  Winter}]{caption14}
\bibinfo{author}{S.~Nikiforova}, \bibinfo{author}{T.~Deoskar},
  \bibinfo{author}{D.~Paperno}, \bibinfo{author}{Y.~Winter},
\newblock \bibinfo{title}{Generating image captions with external encyclopedic
  knowledge},
\newblock \bibinfo{journal}{ArXiv} \bibinfo{volume}{abs/2210.04806}
  (\bibinfo{year}{2022}).
\bibitem[{Xia et~al.(2020)Xia, Huang, Duan, Zhang, Ji, Sui, Cui, Bharti, Liu,
  and Zhou}]{xgpt}
\bibinfo{author}{Q.~Xia}, \bibinfo{author}{H.~Huang},
  \bibinfo{author}{N.~Duan}, \bibinfo{author}{D.~Zhang},
  \bibinfo{author}{L.~Ji}, \bibinfo{author}{Z.~Sui}, \bibinfo{author}{E.~Cui},
  \bibinfo{author}{T.~Bharti}, \bibinfo{author}{X.~Liu},
  \bibinfo{author}{M.~Zhou}, \bibinfo{title}{Xgpt: Cross-modal generative
  pre-training for image captioning}, \bibinfo{year}{2020}. \URLprefix
  \url{https://arxiv.org/abs/2003.01473}.
  \DOIprefix\doi{10.48550/ARXIV.2003.01473}.
\bibitem[{Mokady et~al.(2021)Mokady, Hertz, and Bermano}]{clipcap}
\bibinfo{author}{R.~Mokady}, \bibinfo{author}{A.~Hertz}, \bibinfo{author}{A.~H.
  Bermano},
\newblock \bibinfo{title}{Clipcap: Clip prefix for image captioning},
\newblock \bibinfo{journal}{ArXiv} \bibinfo{volume}{abs/2111.09734}
  (\bibinfo{year}{2021}).
\bibitem[{Luo et~al.(2022)Luo, Xi, Zhang, and Ma}]{caption12}
\bibinfo{author}{Z.~Luo}, \bibinfo{author}{Y.~Xi}, \bibinfo{author}{R.~Zhang},
  \bibinfo{author}{J.~Ma},
\newblock \bibinfo{title}{A frustratingly simple approach for end-to-end image
  captioning},
\newblock \bibinfo{year}{2022}.
\bibitem[{Ramos et~al.(2022)Ramos, Martins, Elliott, and
  Kementchedjhieva}]{caption13}
\bibinfo{author}{R.~P. Ramos}, \bibinfo{author}{B.~Martins},
  \bibinfo{author}{D.~Elliott}, \bibinfo{author}{Y.~Kementchedjhieva},
\newblock \bibinfo{title}{Smallcap: Lightweight image captioning prompted with
  retrieval augmentation},
\newblock \bibinfo{journal}{ArXiv} \bibinfo{volume}{abs/2209.15323}
  (\bibinfo{year}{2022}).
\bibitem[{Luo et~al.(2022)Luo, Hu, Xi, Zhang, and Ma}]{caption11}
\bibinfo{author}{Z.~Luo}, \bibinfo{author}{Z.~Hu}, \bibinfo{author}{Y.~Xi},
  \bibinfo{author}{R.~Zhang}, \bibinfo{author}{J.~Ma},
  \bibinfo{title}{I-tuning: Tuning frozen language models with image for
  lightweight image captioning}, \bibinfo{year}{2022}. \URLprefix
  \url{https://arxiv.org/abs/2202.06574}.
  \DOIprefix\doi{10.48550/ARXIV.2202.06574}.
\bibitem[{Lin et~al.(2014)Lin, Maire, Belongie, Hays, Perona, Ramanan,
  Doll{\'a}r, and Zitnick}]{coco}
\bibinfo{author}{T.-Y. Lin}, \bibinfo{author}{M.~Maire},
  \bibinfo{author}{S.~Belongie}, \bibinfo{author}{J.~Hays},
  \bibinfo{author}{P.~Perona}, \bibinfo{author}{D.~Ramanan},
  \bibinfo{author}{P.~Doll{\'a}r}, \bibinfo{author}{C.~L. Zitnick},
\newblock \bibinfo{title}{Microsoft coco: Common objects in context},
\newblock in: \bibinfo{editor}{D.~Fleet}, \bibinfo{editor}{T.~Pajdla},
  \bibinfo{editor}{B.~Schiele}, \bibinfo{editor}{T.~Tuytelaars} (Eds.),
  \bibinfo{booktitle}{Computer Vision -- ECCV 2014},
  \bibinfo{publisher}{Springer International Publishing},
  \bibinfo{address}{Cham}, \bibinfo{year}{2014}, pp. \bibinfo{pages}{740--755}.
\bibitem[{Young et~al.(2014)Young, Lai, Hodosh, and Hockenmaier}]{flickr}
\bibinfo{author}{P.~Young}, \bibinfo{author}{A.~Lai},
  \bibinfo{author}{M.~Hodosh}, \bibinfo{author}{J.~Hockenmaier},
\newblock \bibinfo{title}{From image descriptions to visual denotations: New
  similarity metrics for semantic inference over event descriptions},
\newblock \bibinfo{journal}{Transactions of the Association for Computational
  Linguistics} \bibinfo{volume}{2} (\bibinfo{year}{2014})
  \bibinfo{pages}{67--78}. \URLprefix \url{https://aclanthology.org/Q14-1006}.
  \DOIprefix\doi{10.1162/tacl_a_00166}.
\bibitem[{Yang et~al.(2019)Yang, Luo, Chen, Li, Yin, He, and Sun}]{story1}
\bibinfo{author}{P.~Yang}, \bibinfo{author}{F.~Luo}, \bibinfo{author}{P.~Chen},
  \bibinfo{author}{L.~Li}, \bibinfo{author}{Z.~Yin}, \bibinfo{author}{X.~He},
  \bibinfo{author}{X.~Sun},
\newblock \bibinfo{title}{Knowledgeable storyteller: A commonsense-driven
  generative model for visual storytelling},
\newblock in: \bibinfo{booktitle}{Proceedings of the Twenty-Eighth
  International Joint Conference on Artificial Intelligence, {IJCAI-19}},
  \bibinfo{publisher}{International Joint Conferences on Artificial
  Intelligence Organization}, \bibinfo{year}{2019}, pp.
  \bibinfo{pages}{5356--5362}. \URLprefix
  \url{https://doi.org/10.24963/ijcai.2019/744}.
  \DOIprefix\doi{10.24963/ijcai.2019/744}.
\bibitem[{Hsu et~al.(2019)Hsu, Chen, Hsu, Li, Lin, Huang, and Ku}]{story2}
\bibinfo{author}{C.-C. Hsu}, \bibinfo{author}{Z.-Y. Chen},
  \bibinfo{author}{C.-Y. Hsu}, \bibinfo{author}{C.-C. Li},
  \bibinfo{author}{T.-Y. Lin}, \bibinfo{author}{T.-H.~K. Huang},
  \bibinfo{author}{L.-W. Ku}, \bibinfo{title}{Knowledge-enriched visual
  storytelling}, \bibinfo{year}{2019}.
  \href{http://arxiv.org/abs/1912.01496}{{\tt arXiv:1912.01496}}.
\bibitem[{Xu et~al.(2021)Xu, Yang, Li, Shen, Ao, and Xu}]{story3}
\bibinfo{author}{C.~Xu}, \bibinfo{author}{M.~Yang}, \bibinfo{author}{C.~Li},
  \bibinfo{author}{Y.~Shen}, \bibinfo{author}{X.~Ao}, \bibinfo{author}{R.~Xu},
\newblock \bibinfo{title}{Imagine, reason and write: Visual storytelling with
  graph knowledge and relational reasoning},
\newblock in: \bibinfo{booktitle}{AAAI}, \bibinfo{year}{2021}.
\bibitem[{Chen et~al.(2021)Chen, Huang, Takamura, and Nakayama}]{story4}
\bibinfo{author}{H.~Chen}, \bibinfo{author}{Y.~Huang},
  \bibinfo{author}{H.~Takamura}, \bibinfo{author}{H.~Nakayama},
\newblock \bibinfo{title}{Commonsense knowledge aware concept selection for
  diverse and informative visual storytelling},
\newblock in: \bibinfo{booktitle}{AAAI}, \bibinfo{year}{2021}.
\bibitem[{Li et~al.(2019)Li, Gan, Shen, Liu, Cheng, Wu, Carin, Carlson, and
  Gao}]{storygan}
\bibinfo{author}{Y.~Li}, \bibinfo{author}{Z.~Gan}, \bibinfo{author}{Y.~Shen},
  \bibinfo{author}{J.~Liu}, \bibinfo{author}{Y.~Cheng},
  \bibinfo{author}{Y.~Wu}, \bibinfo{author}{L.~Carin},
  \bibinfo{author}{D.~Carlson}, \bibinfo{author}{J.~Gao},
\newblock \bibinfo{title}{Storygan: A sequential conditional gan for story
  visualization},
\newblock \bibinfo{year}{2019}, pp. \bibinfo{pages}{6322--6331}.
  \DOIprefix\doi{10.1109/CVPR.2019.00649}.
\bibitem[{Maharana et~al.(2021)Maharana, Hannan, and
  Bansal}]{Maharana2021ImprovingGA}
\bibinfo{author}{A.~Maharana}, \bibinfo{author}{D.~Hannan},
  \bibinfo{author}{M.~Bansal},
\newblock \bibinfo{title}{Improving generation and evaluation of visual stories
  via semantic consistency},
\newblock \bibinfo{journal}{ArXiv} \bibinfo{volume}{abs/2105.10026}
  (\bibinfo{year}{2021}).
\bibitem[{Maharana and Bansal(2021)}]{Maharana2021IntegratingVL}
\bibinfo{author}{A.~Maharana}, \bibinfo{author}{M.~Bansal},
\newblock \bibinfo{title}{Integrating visuospatial, linguistic, and commonsense
  structure into story visualization},
\newblock \bibinfo{journal}{ArXiv} \bibinfo{volume}{abs/2110.10834}
  (\bibinfo{year}{2021}).
\bibitem[{Tsakas et~al.(2023)Tsakas, Lymperaiou, Filandrianos, and
  Stamou}]{impartial}
\bibinfo{author}{N.~Tsakas}, \bibinfo{author}{M.~Lymperaiou},
  \bibinfo{author}{G.~Filandrianos}, \bibinfo{author}{G.~Stamou},
  \bibinfo{title}{An impartial transformer for story visualization},
  \bibinfo{year}{2023}. \URLprefix \url{https://arxiv.org/abs/2301.03563}.
  \DOIprefix\doi{10.48550/ARXIV.2301.03563}.
\bibitem[{Goodfellow et~al.(2014)Goodfellow, Pouget-Abadie, Mirza, Xu,
  Warde-Farley, Ozair, Courville, and Bengio}]{gan}
\bibinfo{author}{I.~Goodfellow}, \bibinfo{author}{J.~Pouget-Abadie},
  \bibinfo{author}{M.~Mirza}, \bibinfo{author}{B.~Xu},
  \bibinfo{author}{D.~Warde-Farley}, \bibinfo{author}{S.~Ozair},
  \bibinfo{author}{A.~Courville}, \bibinfo{author}{Y.~Bengio},
\newblock \bibinfo{title}{Generative adversarial nets},
\newblock in: \bibinfo{booktitle}{Advances in Neural Information Processing
  Systems}, volume~\bibinfo{volume}{27}, \bibinfo{year}{2014}. \URLprefix
  \url{https://proceedings.neurips.cc/paper/2014/file/5ca3e9b122f61f8f06494c97b1afccf3-Paper.pdf}.
\bibitem[{Maharana et~al.(2022)Maharana, Hannan, and Bansal}]{storydalle}
\bibinfo{author}{A.~Maharana}, \bibinfo{author}{D.~Hannan},
  \bibinfo{author}{M.~Bansal}, \bibinfo{title}{Storydall-e: Adapting pretrained
  text-to-image transformers for story continuation}, \bibinfo{year}{2022}.
  \URLprefix \url{https://arxiv.org/abs/2209.06192}.
  \DOIprefix\doi{10.48550/ARXIV.2209.06192}.
\bibitem[{Ramesh et~al.(2021)Ramesh, Pavlov, Goh, Gray, Voss, Radford, Chen,
  and Sutskever}]{dalle}
\bibinfo{author}{A.~Ramesh}, \bibinfo{author}{M.~Pavlov},
  \bibinfo{author}{G.~Goh}, \bibinfo{author}{S.~Gray},
  \bibinfo{author}{C.~Voss}, \bibinfo{author}{A.~Radford},
  \bibinfo{author}{M.~Chen}, \bibinfo{author}{I.~Sutskever},
  \bibinfo{title}{Zero-shot text-to-image generation}, \bibinfo{year}{2021}.
  \href{http://arxiv.org/abs/2102.12092}{{\tt arXiv:2102.12092}}.
\bibitem[{Sharma et~al.(2018)Sharma, Ding, Goodman, and Soricut}]{cc}
\bibinfo{author}{P.~Sharma}, \bibinfo{author}{N.~Ding},
  \bibinfo{author}{S.~Goodman}, \bibinfo{author}{R.~Soricut},
\newblock \bibinfo{title}{Conceptual captions: A cleaned, hypernymed, image
  alt-text dataset for automatic image captioning},
\newblock in: \bibinfo{booktitle}{ACL}, \bibinfo{year}{2018}.
\bibitem[{Ordonez et~al.(2011)Ordonez, Kulkarni, and Berg}]{sbu}
\bibinfo{author}{V.~Ordonez}, \bibinfo{author}{G.~Kulkarni},
  \bibinfo{author}{T.~Berg},
\newblock \bibinfo{title}{Im2text: Describing images using 1 million captioned
  photographs},
\newblock in: \bibinfo{editor}{J.~Shawe-Taylor}, \bibinfo{editor}{R.~Zemel},
  \bibinfo{editor}{P.~Bartlett}, \bibinfo{editor}{F.~Pereira},
  \bibinfo{editor}{K.~Weinberger} (Eds.), \bibinfo{booktitle}{Advances in
  Neural Information Processing Systems}, volume~\bibinfo{volume}{24},
  \bibinfo{publisher}{Curran Associates, Inc.}, \bibinfo{year}{2011}.
  \URLprefix
  \url{https://proceedings.neurips.cc/paper/2011/file/5dd9db5e033da9c6fb5ba83c7a7ebea9-Paper.pdf}.
\bibitem[{Marasovi{\'c} et~al.(2020)Marasovi{\'c}, Bhagavatula, Park, Bras,
  Smith, and Choi}]{multi2}
\bibinfo{author}{A.~Marasovi{\'c}}, \bibinfo{author}{C.~Bhagavatula},
  \bibinfo{author}{J.~S. Park}, \bibinfo{author}{R.~L. Bras},
  \bibinfo{author}{N.~A. Smith}, \bibinfo{author}{Y.~Choi},
\newblock \bibinfo{title}{Natural language rationales with full-stack visual
  reasoning: From pixels to semantic frames to commonsense graphs},
\newblock in: \bibinfo{booktitle}{FINDINGS}, \bibinfo{year}{2020}.
\bibitem[{Shevchenko et~al.(2021)Shevchenko, Teney, Dick, and van~den
  Hengel}]{multi3}
\bibinfo{author}{V.~Shevchenko}, \bibinfo{author}{D.~Teney},
  \bibinfo{author}{A.~R. Dick}, \bibinfo{author}{A.~van~den Hengel},
\newblock \bibinfo{title}{Reasoning over vision and language: Exploring the
  benefits of supplemental knowledge},
\newblock \bibinfo{journal}{ArXiv} \bibinfo{volume}{abs/2101.06013}
  (\bibinfo{year}{2021}).
\bibitem[{Chen et~al.(2021)Chen, Huang, Bisk, McDuff, and Gao}]{multi6}
\bibinfo{author}{K.~Chen}, \bibinfo{author}{Q.~Huang},
  \bibinfo{author}{Y.~Bisk}, \bibinfo{author}{D.~J. McDuff},
  \bibinfo{author}{J.~Gao},
\newblock \bibinfo{title}{Kb-vlp: Knowledge based vision and language
  pretraining},
\newblock \bibinfo{year}{2021}.
\bibitem[{Hu et~al.(2022)Hu, Hua, Yang, Shi, Smith, and Luo}]{promptcap}
\bibinfo{author}{Y.~Hu}, \bibinfo{author}{H.~Hua}, \bibinfo{author}{Z.~Yang},
  \bibinfo{author}{W.~Shi}, \bibinfo{author}{N.~A. Smith},
  \bibinfo{author}{J.~Luo},
\newblock \bibinfo{title}{Promptcap: Prompt-guided task-aware image
  captioning},
\newblock \bibinfo{journal}{ArXiv} \bibinfo{volume}{abs/2211.09699}
  (\bibinfo{year}{2022}).
\bibitem[{Zeng et~al.(2022)Zeng, Wong, Welker, Choromanski, Tombari, Purohit,
  Ryoo, Sindhwani, Lee, Vanhoucke, and Florence}]{socratic}
\bibinfo{author}{A.~Zeng}, \bibinfo{author}{A.~S. Wong},
  \bibinfo{author}{S.~Welker}, \bibinfo{author}{K.~Choromanski},
  \bibinfo{author}{F.~Tombari}, \bibinfo{author}{A.~Purohit},
  \bibinfo{author}{M.~S. Ryoo}, \bibinfo{author}{V.~Sindhwani},
  \bibinfo{author}{J.~Lee}, \bibinfo{author}{V.~Vanhoucke},
  \bibinfo{author}{P.~R. Florence},
\newblock \bibinfo{title}{Socratic models: Composing zero-shot multimodal
  reasoning with language},
\newblock \bibinfo{journal}{ArXiv} \bibinfo{volume}{abs/2204.00598}
  (\bibinfo{year}{2022}).
\bibitem[{Geva et~al.(2021)Geva, Khashabi, Segal, Khot, Roth, and
  Berant}]{implicit-strategy}
\bibinfo{author}{M.~Geva}, \bibinfo{author}{D.~Khashabi},
  \bibinfo{author}{E.~Segal}, \bibinfo{author}{T.~Khot},
  \bibinfo{author}{D.~Roth}, \bibinfo{author}{J.~Berant},
\newblock \bibinfo{title}{{Did Aristotle Use a Laptop? A Question Answering
  Benchmark with Implicit Reasoning Strategies}},
\newblock \bibinfo{journal}{Transactions of the Association for Computational
  Linguistics} \bibinfo{volume}{9} (\bibinfo{year}{2021})
  \bibinfo{pages}{346--361}. \URLprefix
  \url{https://doi.org/10.1162/tacl\_a\_00370}.
  \DOIprefix\doi{10.1162/tacl_a_00370}.
\bibitem[{Talmor et~al.(2019)Talmor, Herzig, Lourie, and
  Berant}]{commonsenseqa}
\bibinfo{author}{A.~Talmor}, \bibinfo{author}{J.~Herzig},
  \bibinfo{author}{N.~Lourie}, \bibinfo{author}{J.~Berant},
\newblock \bibinfo{title}{{C}ommonsense{QA}: A question answering challenge
  targeting commonsense knowledge},
\newblock in: \bibinfo{booktitle}{Proceedings of the 2019 Conference of the
  North {A}merican Chapter of the Association for Computational Linguistics:
  Human Language Technologies, Volume 1 (Long and Short Papers)},
  \bibinfo{publisher}{Association for Computational Linguistics},
  \bibinfo{address}{Minneapolis, Minnesota}, \bibinfo{year}{2019}, pp.
  \bibinfo{pages}{4149--4158}. \URLprefix
  \url{https://aclanthology.org/N19-1421}.
  \DOIprefix\doi{10.18653/v1/N19-1421}.
\bibitem[{Srivastava et~al.(2022)Srivastava, Rastogi, Rao, Shoeb, Abid, Fisch,
  Brown, Santoro, Gupta, Garriga-Alonso, Kluska, Lewkowycz, Agarwal, Power,
  Ray, Warstadt, Kocurek, Safaya, Tazarv, and Wu}]{imitation-game}
\bibinfo{author}{A.~Srivastava}, \bibinfo{author}{A.~Rastogi},
  \bibinfo{author}{A.~Rao}, \bibinfo{author}{A.~Shoeb},
  \bibinfo{author}{A.~Abid}, \bibinfo{author}{A.~Fisch},
  \bibinfo{author}{A.~Brown}, \bibinfo{author}{A.~Santoro},
  \bibinfo{author}{A.~Gupta}, \bibinfo{author}{A.~Garriga-Alonso},
  \bibinfo{author}{A.~Kluska}, \bibinfo{author}{A.~Lewkowycz},
  \bibinfo{author}{A.~Agarwal}, \bibinfo{author}{A.~Power},
  \bibinfo{author}{A.~Ray}, \bibinfo{author}{A.~Warstadt},
  \bibinfo{author}{A.~Kocurek}, \bibinfo{author}{A.~Safaya},
  \bibinfo{author}{A.~Tazarv}, \bibinfo{author}{Z.~Wu}, \bibinfo{title}{Beyond
  the imitation game: Quantifying and extrapolating the capabilities of
  language models}, \bibinfo{year}{2022}.
  \DOIprefix\doi{10.48550/arXiv.2206.04615}.
\bibitem[{Islam et~al.(2021)Islam, Eberle, Ghafoor, and Ahmed}]{xai-survey}
\bibinfo{author}{S.~R. Islam}, \bibinfo{author}{W.~Eberle},
  \bibinfo{author}{S.~K. Ghafoor}, \bibinfo{author}{M.~Ahmed},
  \bibinfo{title}{Explainable artificial intelligence approaches: A survey},
  \bibinfo{year}{2021}. \URLprefix \url{https://arxiv.org/abs/2101.09429}.
  \DOIprefix\doi{10.48550/ARXIV.2101.09429}.
\bibitem[{Yamada et~al.(2020)Yamada, Asai, Shindo, Takeda, and
  Matsumoto}]{luke}
\bibinfo{author}{I.~Yamada}, \bibinfo{author}{A.~Asai},
  \bibinfo{author}{H.~Shindo}, \bibinfo{author}{H.~Takeda},
  \bibinfo{author}{Y.~Matsumoto},
\newblock \bibinfo{title}{Luke: Deep contextualized entity representations with
  entity-aware self-attention},
\newblock in: \bibinfo{booktitle}{EMNLP}, \bibinfo{year}{2020}.
\bibitem[{Qin et~al.(2021)Qin, Lin, Takanobu, Liu, Li, Ji, Huang, Sun, and
  Zhou}]{erica}
\bibinfo{author}{Y.~Qin}, \bibinfo{author}{Y.~Lin},
  \bibinfo{author}{R.~Takanobu}, \bibinfo{author}{Z.~Liu},
  \bibinfo{author}{P.~Li}, \bibinfo{author}{H.~Ji}, \bibinfo{author}{M.~Huang},
  \bibinfo{author}{M.~Sun}, \bibinfo{author}{J.~Zhou},
\newblock \bibinfo{title}{Erica: Improving entity and relation understanding
  for pre-trained language models via contrastive learning},
\newblock in: \bibinfo{booktitle}{ACL/IJCNLP}, \bibinfo{year}{2021}.
\bibitem[{Poerner et~al.(2020)Poerner, Waltinger, and Sch{\"u}tze}]{ebert}
\bibinfo{author}{N.~Poerner}, \bibinfo{author}{U.~Waltinger},
  \bibinfo{author}{H.~Sch{\"u}tze},
\newblock \bibinfo{title}{{E}-{BERT}: Efficient-yet-effective entity embeddings
  for {BERT}},
\newblock in: \bibinfo{booktitle}{Findings of the Association for Computational
  Linguistics: EMNLP 2020}, \bibinfo{publisher}{Association for Computational
  Linguistics}, \bibinfo{address}{Online}, \bibinfo{year}{2020}, pp.
  \bibinfo{pages}{803--818}. \URLprefix
  \url{https://aclanthology.org/2020.findings-emnlp.71}.
  \DOIprefix\doi{10.18653/v1/2020.findings-emnlp.71}.
\bibitem[{Bauer et~al.(2021)Bauer, Deng, and Bansal}]{ernie-nli}
\bibinfo{author}{L.~Bauer}, \bibinfo{author}{L.~Deng},
  \bibinfo{author}{M.~Bansal},
\newblock \bibinfo{title}{Ernie-nli: Analyzing the impact of domain-specific
  external knowledge on enhanced representations for nli},
\newblock in: \bibinfo{booktitle}{DEELIO}, \bibinfo{year}{2021}.
\bibitem[{Sharir et~al.(2020)Sharir, Peleg, and Shoham}]{pretraining-cost}
\bibinfo{author}{O.~Sharir}, \bibinfo{author}{B.~Peleg},
  \bibinfo{author}{Y.~Shoham}, \bibinfo{title}{The cost of training nlp models:
  A concise overview}, \bibinfo{year}{2020}. \URLprefix
  \url{https://arxiv.org/abs/2004.08900}.
  \DOIprefix\doi{10.48550/ARXIV.2004.08900}.
\bibitem[{Floridi and Chiriatti(2020)}]{gpt3-kraksimo}
\bibinfo{author}{L.~Floridi}, \bibinfo{author}{M.~Chiriatti},
\newblock \bibinfo{title}{Gpt-3: Its nature, scope, limits, and consequences},
\newblock \bibinfo{journal}{Minds and Machines} \bibinfo{volume}{30}
  (\bibinfo{year}{2020}) \bibinfo{pages}{681--694}.
\bibitem[{Kassner and Schütze(2019)}]{factcheck}
\bibinfo{author}{N.~Kassner}, \bibinfo{author}{H.~Schütze},
  \bibinfo{title}{Negated and misprimed probes for pretrained language models:
  Birds can talk, but cannot fly}, \bibinfo{year}{2019}. \URLprefix
  \url{https://arxiv.org/abs/1911.03343}.
  \DOIprefix\doi{10.48550/ARXIV.1911.03343}.
\bibitem[{Shi et~al.(2023)Shi, Chen, Misra, Scales, Dohan, hsin Chi, Scharli,
  and Zhou}]{distracted}
\bibinfo{author}{F.~Shi}, \bibinfo{author}{X.~Chen},
  \bibinfo{author}{K.~Misra}, \bibinfo{author}{N.~Scales},
  \bibinfo{author}{D.~Dohan}, \bibinfo{author}{E.~H. hsin Chi},
  \bibinfo{author}{N.~Scharli}, \bibinfo{author}{D.~Zhou},
\newblock \bibinfo{title}{Large language models can be easily distracted by
  irrelevant context},
\newblock \bibinfo{journal}{ArXiv} \bibinfo{volume}{abs/2302.00093}
  (\bibinfo{year}{2023}).
\bibitem[{Yuan et~al.(2022)Yuan, Hu, Vulic, Korhonen, and Meng}]{deductive}
\bibinfo{author}{Z.~Yuan}, \bibinfo{author}{S.~Hu}, \bibinfo{author}{I.~Vulic},
  \bibinfo{author}{A.~Korhonen}, \bibinfo{author}{Z.~Meng},
\newblock \bibinfo{title}{Can pretrained language models (yet) reason
  deductively?},
\newblock \bibinfo{journal}{ArXiv} \bibinfo{volume}{abs/2210.06442}
  (\bibinfo{year}{2022}).
\bibitem[{Kauf et~al.(2022)Kauf, Ivanova, Rambelli, Chersoni, She, Chowdhury,
  Fedorenko, and Lenci}]{unlikely}
\bibinfo{author}{C.~Kauf}, \bibinfo{author}{A.~A. Ivanova},
  \bibinfo{author}{G.~Rambelli}, \bibinfo{author}{E.~Chersoni},
  \bibinfo{author}{J.~S. She}, \bibinfo{author}{Z.~Chowdhury},
  \bibinfo{author}{E.~Fedorenko}, \bibinfo{author}{A.~Lenci},
\newblock \bibinfo{title}{Event knowledge in large language models: the gap
  between the impossible and the unlikely},
\newblock \bibinfo{journal}{ArXiv} \bibinfo{volume}{abs/2212.01488}
  (\bibinfo{year}{2022}).

\end{thebibliography}

\appendix

\end{document}